\documentclass[10pt,twocolumn,letterpaper]{article}

\usepackage[pagenumbers]{cvpr} % To force page numbers, e.g. for an arXiv version

\definecolor{cvprblue}{rgb}{0.21,0.49,0.74}
\usepackage[pagebackref,breaklinks,colorlinks,allcolors=cvprblue]{hyperref}
\usepackage{animate}

\def\paperID{3826} % *** Enter the Paper ID here
\def\confName{CVPR}
\def\confYear{2025}

\newcommand{\cherry}[1]{{\color{black} #1}}
\newcommand{\mt}[1]{\noindent {\color{black} #1}}

\newcommand{\modelname}[1]{\noindent{MotionBridge}}
\definecolor{darkg}{RGB}{0, 150, 0}
\newcommand{\simon}[1]{{\color{black} #1}}

\begin{document}

\title{\cherry{\modelname{}: Dynamic Video Inbetweening with Flexible Controls}}

\author{%
    Maham Tanveer$^{1,2}\thanks{Work done as intern at Adobe.}$ \quad Yang Zhou$^2$ \quad Simon Niklaus$^2$ \\
    Ali Mahdavi Amiri$^1$ \quad Hao Zhang$^1$ \quad Krishna Kumar Singh$^2$ \quad Nanxuan Zhao$^2$ \\
    $^1$Simon Fraser University \qquad $^2$Adobe Research\\ 
}

% \maketitle

% \twocolumn[{
% % \renewcommand
% % \twocolumn[1][]{#1}%
% \maketitle
% \begin{center}
%     \centering
%     % \captionsetup{type=figure}
%     \includegraphics[width=1\textwidth]{data/teaser_small.pdf}
%     \captionof{figure}{
%     \cherry{\modelname{} generates smooth and plausible transitions between two RGB image frames following user-specified trajectories, producing large and intricate motions (see top two rows for diverse results for the same dog). It also supports multi-object control where the motions vary between objects (the bee + flowers), as well as mask control specifying static (red mask) vs.~dynamic regions (blue mask); see last two rows. In the last row, the static mask helps maintain the lady in the same position while turning her body naturally.}
%     % \mt{At the top, we show two results with the same start and end images but different in-between trajectories. The second row shows a multi object movement, with the bee in a specific trajectory and the flowers being swayed. Lastly, a detailed masked result is shown where red mask shows static and blue mask shows dynamic, showing how to use masks to control regions of movement.
%     % } 
%     % \todoc{overlay, last example, mention secs?, gif, color}
%     }
%     \label{fig:teaser}
% \end{center}%
% }]

\twocolumn[{
\maketitle
\begin{center}
    \centering
    \includegraphics[width=1\textwidth]{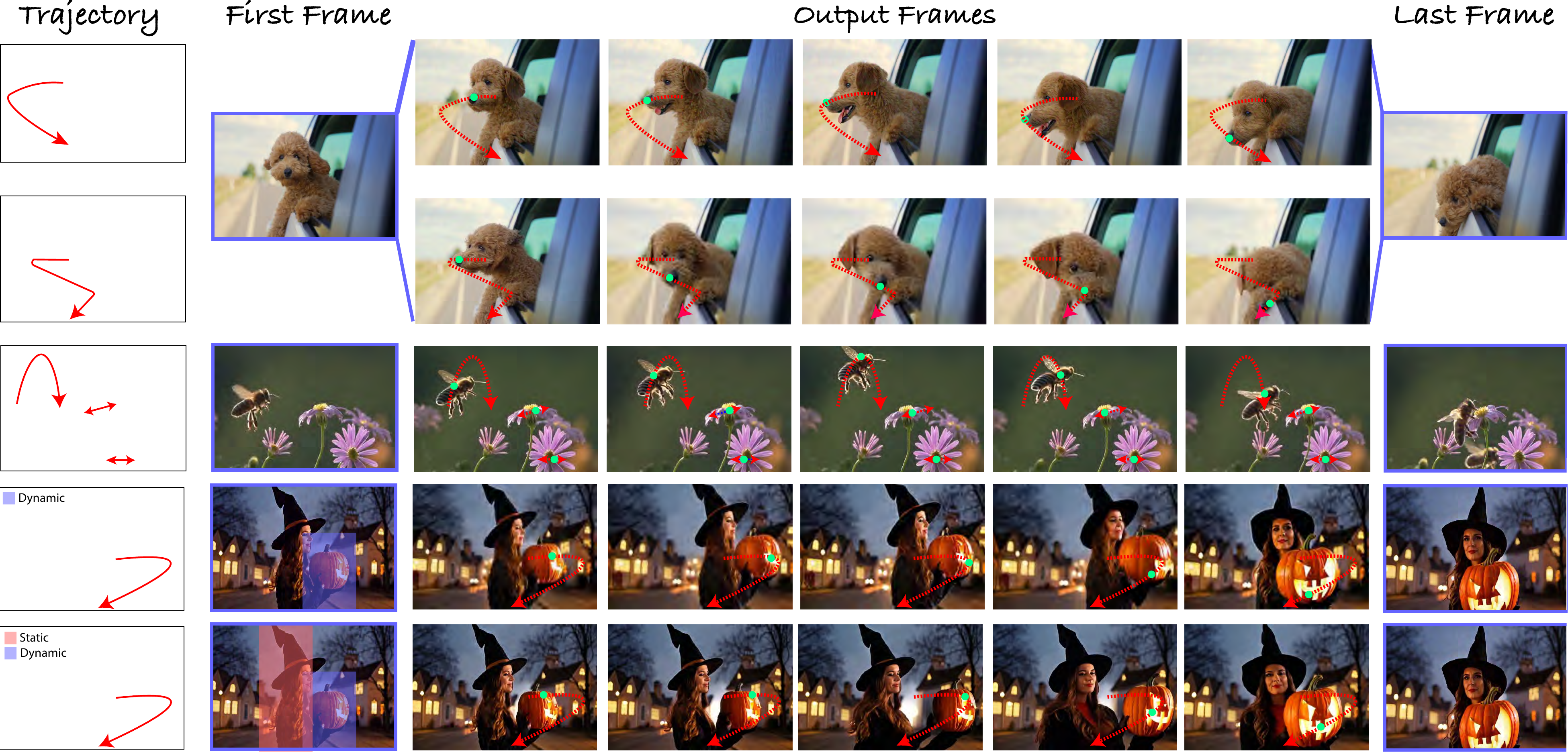}
    \captionof{figure}{
    \modelname{} generates smooth and plausible transitions between two RGB images following user-defined trajectories, producing large and intricate motions (see top two rows for diverse results for the same dog). It offers multi-object control with motions varying between objects (bee + flowers), as well as mask control specifying static (red) vs.~dynamic regions (blue); see last two rows. In last row, the static mask helps maintain the lady in the same position while turning her body naturally.
    }
    \label{fig:teaser}
\end{center}
}]

\let\thefootnote\relax\footnotetext{*Work done during an internship at Adobe.}

\begin{abstract}
\cherry{By generating plausible and smooth transitions between two image frames, video inbetweening is an essential tool for video editing and long video synthesis. Traditional works lack the capability to generate complex large motions. 
% \kr{This part is not clear, not sure how it restricts control space}
While recent video generation techniques are powerful in creating high-quality results, they often lack fine control over the details of intermediate frames, which can lead to results that do not align with the creative mind. We introduce \modelname{}, a unified video inbetweening framework that allows flexible controls, including trajectory strokes, keyframes, masks, guide pixels, and text.
However, learning such multi-modal controls in a unified framework is a challenging task. We thus design two generators to extract the control signal faithfully and encode features through dual-branch embedders to resolve ambiguities. We further introduce a curriculum training strategy to smoothly learn various controls.
Extensive qualitative and quantitative experiments have demonstrated that such multi-modal controls enable a more dynamic, customizable, and contextually accurate visual narrative.} Please visit our project website: \href{https://motionbridge.github.io/}{https://motionbridge.github.io/}

% In this project we present a novel approach to controllable inbetweening or interpolation for animation and video generation. By utilizing a diffusion model, we leverage sparse keyframes, motion paths, and region-specific masks to generate interpolated frames that bridge given keyframes seamlessly. The system incorporates various conditions, including full and masked images, to guide the model's output while maintaining flexibility and creative freedom. This method allows for enhanced control over the interpolation process, enabling more dynamic and contextually accurate short animations. 
\end{abstract}    
\section{Introduction}
\label{sec:intro}

\cherry{Video Inbetweening refers to the process of generating intermediate frames between two keyframes, creating a smooth transition from one scene to another. It is becoming an increasingly important building block for video content creators and animators to conduct video editing, storytelling, and short-to-long video synthesis~\cite{meyer2018colorprop, siyao2021animeinterp}. } % As a long-standing problem, traditional methods~\todoc{cites} often rely on \todoc{add basic ideas, optical flow? manual-crafted?} features, which have limited generation power and make it hard to generate enough video frames to accommodate diverse motions.
\simon{Frame interpolation is typically done in two steps, motion estimation and motion compensation~\cite{ha2004moco, choi2000fruc, niklaus2020softsplat, reda2022film}. 
% However, the further apart the input frames are the more difficult such an explicit motion estimation and compensation becomes since we now have to deal with generating new/plausible content to make up for information that is missing from the input.
} 
% \kr{do we mean compensation becomes hard? also we should say 'large' before 'generating new/plausible content}
% This technique is essential for enhancing the fluidity and realism of animations, as it allows for more natural motion without the need for manual frame-by-frame adjustments. By enabling smoother transitions, interpolation reduces the workload on animators and can be particularly useful for short animations, gaming, and video editing. With advancements in AI and machine learning, modern interpolation techniques can now incorporate sophisticated guidance, such as motion paths and selective masking, which opens up new possibilities for creative storytelling and streamlined production workflows.
\cherry{However, as the temporal or spatial gap between input frames widens, motion estimation and compensation become increasingly difficult, as generating realistic intermediate frames requires synthesizing novel content to bridge the missing information between inputs. With the emerging success of video generative models, the exploration space for the generated frames becomes larger}\simon{, thus opening up new possibilities for inbetweening of distant inputs. At the same time, just blindly applying a foundational video model will typically not suffice since users are often not interested in just a possible interpolation result but one that follows their artistic expression through various means of control.} % Simply relying on texts and images often fails to generate satisfactory results.}

\cherry{To this end, we propose \modelname{}, the initial effort for conducting controllable video inbetweening, which can generate diverse large motions through multi-modal controls (\eg, trajectories, keyframes, masks, guide pixels, and text), in a unified framework. This allows users to generate dynamic, accurate, and customizable results. We take advantage of the Diffusion Transformer (DiT) architecture~\cite{peebles2023scalable}, which shows promising capability for generating long and high-quality videos. We design our model in a backbone-agnostic manner,
% plug-and-play manner, 
which can work with different DiT designs/backbones.}

% \mt{\modelname{} introduces the first method for performing controllable inbetweening, combining both input images and motion control in a unified framework. Our primary contribution is an effective method for integrating motion and image content into the DiT video diffusion model within an interpolation framework. This plug-and-play system can work with different base models. For our purpose we use a two-branch system, where one branch deals with the content or image information and the other with motion input.

\cherry{Technically, our model is characterized by several core design choices to address the unique challenges of our task. 1). Rather than fusing all the control signals together all at once, to reduce ambiguity, we group controls into two categories: content control (\eg, masks and guide pixels) and motion controls (\eg, trajectories). We then utilize dual-branch embedders to compute the required features respectively before guiding the denoising process. 
2). Representing video motion control with simple yet accurate representations is challenging. We propose a generator that synthesizes trajectories from optical flow and converts them into sparse RGB points as the motion representation used in model training.
% Extracting the sparse trajectories from video frames while ensuring both accuracy and simplicity is a challenging task. We design a generator to synthesize trajectories from optical flow, and then convert them into sparse RGB points as an intermediate representation for training.  
%
3). We go beyond conventional trajectory control by complementing it with spatial content control such as masks and guide pixels. Through these, users can specify the regions they want to move or keep static.
% \kr{we need to say more exactly what it controls like object/region we want to move or not move}, and propose an Augmented Frame Generator. This module can generate controls based on keyframes and extracted motion trajectories. \kr{the motivation of augment 
%  frame generator is not clear, we don't need to go in details about how it does, but give high level intuition for it} 
 It helps further reduce ambiguity in the generation, offering a soft condition, as shown in the last example of Fig.~\ref{fig:teaser} (last two rows). 
4). With multi-modal controls, straightforward training does not work well, and we thus propose a curriculum learning strategy to ensure the model learns various controls smoothly. We feed the model with more dense and easy control, and gradually move to more sparse and high-level control.
 % \kr{need to briefly (in a sentence) explain for curriculum, how conditioning is introduced gradually and reasoning behind it.} 
 }

% We first convert a text-to-video diffusion model into an image-to-video diffusion model so that it may take multiple input images to guide the output generation. 
% For motion, we use motion trajectories converted to sparse optical flow. We detail how this input can be used as a condition in latest DiT architecture, demonstrating its effectiveness in guiding the generative process, where we also show the effectiveness of base DiT-VAE for embedding motion information.

 % Beyond this, our framework introduces a novel control mechanism of augmented frames. Augmented frames are synthetic frames generated from input keyframes and motion trajectory. It provides a soft conditioning to the model and cues to remove ambiguity in generation of output and  offering additional control over the motion generation. As an example please see bottom row of Fig ~\ref{fig:teaser} where a static and dynamic mask control regions of movement. This enables not only more varied and refined inbetweening but also empowers users with creative flexibility in shaping the temporal flow of generated content. Finally, we show how text prompts provide an additional layer of flexibility. 

\cherry{We conduct extensive experiments to evaluate the effectiveness of \modelname{} both quantitatively and qualitatively. We also demonstrate several practical applications. We found our model is rather powerful and can go beyond the inbetweening scenarios, to work on controllable image-to-video (I2V) generation. Furthermore, our model can not only customize results but also improve the text-to-video (T2V) generation quality by reducing ambiguity.
} 
% We elaborate on the diverse range of controls enabled by this methodology, illustrating how both motion and semantic content can be adjusted independently or in tandem, paving the way for more personalized and expressive inbetweening.
% We outline the contributions of our model:
% \begin{itemize}[label=\textbullet] 
% \item First method to develop controllable inbetweening generation.
% \item A variety of controls including content, motion, masks and text.
% \item Lightweight architecture designed to embed both content and motion information in a two-branch system, offering a ``general-purpose'' plug-and-play technique applicable across models with various transformer backbone architectures.
% \item Uses RGB optical flow to work with the base DiT architecture 
% \item Employs sparse optical flow with DiT-VAE to embed motion effectively.
% \item Demonstrated the effectiveness of the model through rigorous experiments, validating its performance in generating coherent and high-quality video outputs.
% \end{itemize}
\cherry{Our contributions are summarized as below:
\begin{itemize}
    \item We take the initial effort to solve controllable video inbetweening task that supports multi-modal controls for customizing diverse large motions, in a unified framework.
    \item We group controls into two sets (\ie, content and motion) and encode them through dual-branch embedders.
    \item We introduce two separate generators to extract compact control signals, and design a curriculum training strategy to learn multi-control sequentially.
    \item We demonstrate the flexibility and superior performance of our model through extensive experiments.
    % and applications.
\end{itemize}}

\begin{figure*}[t]
\centering
  \includegraphics[width=1\textwidth]{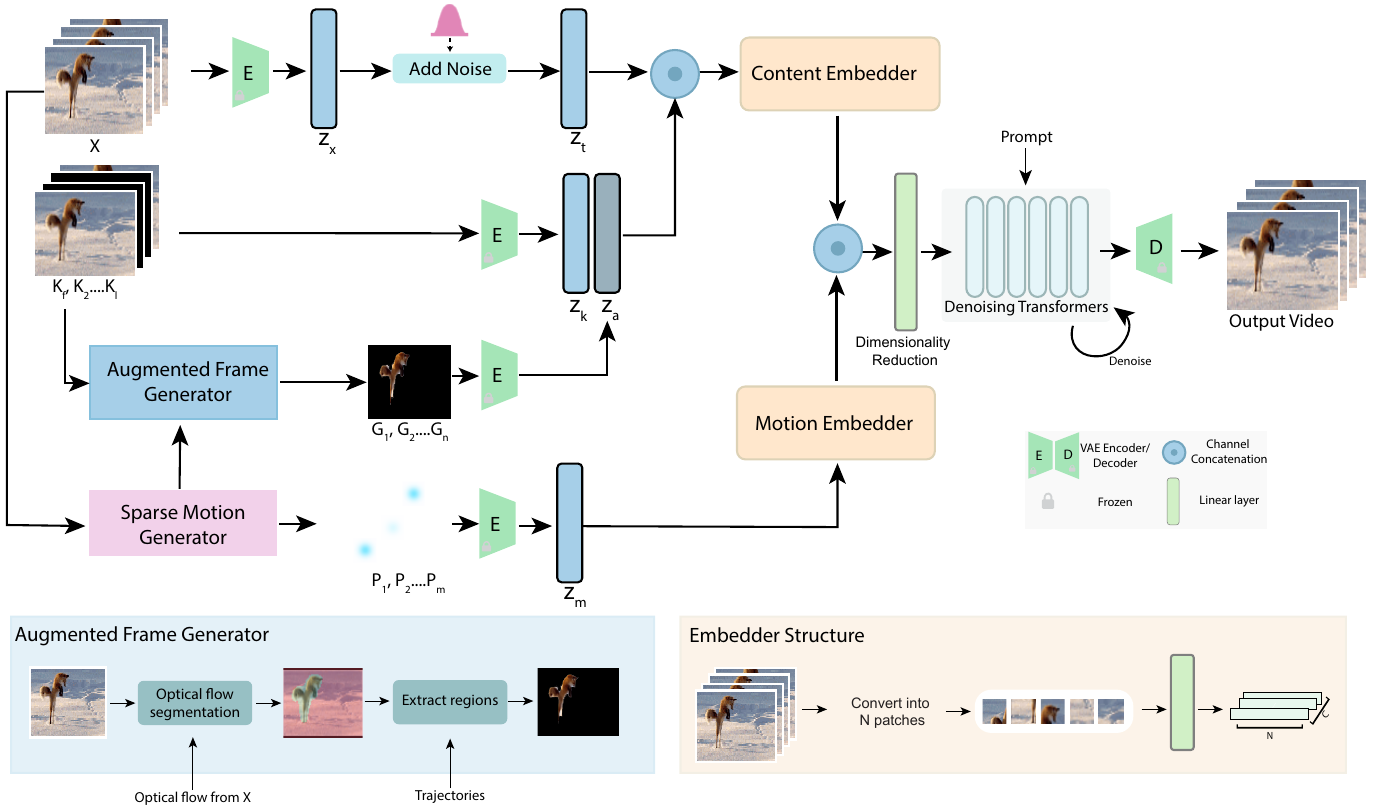}
  \caption{
  \cherry{Overview of our \modelname{} pipeline. Given a video $X$, we propose a  Sparse Motion Generator to provide conditioning for the motion trajectory with sparse RGB point controls, and an Augmented Frame Generator to compute guiding pixels for providing fine-grained control. The control signals are encoded through dual-branch embedders respectively to capture accurate content and motion features. Our model is flexible to take multi-modal controls for interpolation during the inference. }
  % \mt{A video X is processed to extract keyframes and sparse motion paths. VAE is used to compress and encode the input sequences. Keyframes are further enhanced by integrating augmented keyframes based on their content and motion paths. On the right, the ``Embedder" structure for video and motion is shown with video as example input. Inputs are patchified and then embedded. The outputs of video and motion embedder are then concatenated along ``C" axis. A final linear layer adjusts dimensions to match the model's input requirements. }
  % \todoc{frames, diffusion, legend, labels, symbols need to be consistent with main text} 
  % \ali{Can we change the color of embedders to a light yellow? Also make the motion generator a light pink? Also, the arrow to augmented frame generator is weird.}
  }
  \label{fig:pipeline}
\end{figure*}

\section{Related Work}
\label{sec:related}
% \todoc{highlight the killing points for comparing with related works}

\subsection{\cherry{Video Generation}}
% \todoc{introduce text-based, image-based video generation works, gan-based, diffusion-based, dit-based frameworks, highlight we are work for inbetweening which is different from previous works, a sentence for challenging.}

Creating realistic and novel videos has long been an interesting research problem~\cite{yu2023magvit,ranzato2016video}. Earlier studies have employed various generative models including GANs~\cite{yu2023magvit,saito2017temporal,tulyakov2017mocogan,shen2023mostganv} and temporally aware networks such as LSTM or autoregressive models~\cite{srivastava2016unsupervised,yan2021videogpt,hong2022cogvideo}. Recently, inspired by the success of diffusion models in image synthesis, several works have begun to investigate the use of diffusion models for conditional and unconditional video generation~\cite{ho2022video, singer2022make, ho2022imagen, kwon2024harivo}. Stable Video Diffusion~\cite{blattmann2023stable} leverages latent diffusion models~\cite{rombach2022high, blattmann2023align} for generating temporally coherent content. Few-shot video generation is facilitated by methods like Tune-a-video~\cite{wu2023tune}, which fine-tunes pre-trained image diffusion models, while training-free methods~\cite{hong2023large} leverage large language models for generative guidance. Another approach to generating videos in a controllable manner is to use keyframes along with text conditions~\cite{girdhar2023emu, wang2024microcinema, li2023videogen, zeng2024make}, where initial frames are generated to guide subsequent frames, with latent-consistency networks ensuring temporal and visual coherence. However, our approach is different from such keyframe conditioning techniques as we aim to interpolate between two given frames following \cherry{flexible multi-modal controls in a unified framework.}
% a provided trajectory.

% \kr{should we also add that we provide even more richer control than normal image to video model?}

\subsection{Video Inbetweening}
% \todoc{further details for inbetweening works, previous work are lack of control, @Simon}
% Interpolation techniques have been widely studied in computer vision, primarily to improve frame rates and generate smooth transitions between frames. Traditional methods, such as linear interpolation and optical flow-based techniques, estimate pixel movement between frames to create intermediate outputs. Recent advancements incorporate deep learning approaches, particularly convolutional neural networks (CNNs) and generative models, which have demonstrated improved accuracy and realism. However, these methods often struggle with complex motions and may lack control over specific regions within the frames.
\simon{Video inbetweening has many names such as frame interpolation, frame rate up-conversion, or temporal super-resolution. It has a long history, with early approaches operating at a block- instead of a pixel-level due to compute constraints at the time~\cite{choi2000fruc, ha2004moco}. While we have more compute nowadays, the underlying framework of motion estimation and compensation has largely remained the same throughout the years~\cite{niklaus2020softsplat, niklaus2023splatsyn, reda2022film, niklaus2018ctxsyn}. And even with approaches like phase- or kernel-based interpolation~\cite{niklaus2017adaconv, niklaus2017sepconv, meyer2015phasebased,zhou2022audio}, it is fundamentally still about re-synthesizing an in-between frame from what is in the input frames. However, as the inputs become more distant in time and/or space, the inbetweening will require information that is not present so we need to hallucinate it instead. Nowadays we can utilize foundational video models for generating plausible interpolation results~\cite{feng2024explorative}, but users typically aren't interested in just a possible interpolation result but one that follows their artistic expression. This is where motion control comes into the picture, which is the focus of our work.} 
% \kr{which is the focus of our work.}

\subsection{Motion Control}
% \todoc{highlight we can have larger motion, and diverse multi-level control.}
% \todoc{@Maham, revise this section and add cites}
\mt{
A variety of methods have recently been proposed for controllable video generation. Approaches such as MotionDirector~\cite{zhao2025motiondirector}, Customize-A-Video~\cite{ren2024customize} and Tune-a-Video~\cite{wu2023tune} learn motion patterns from a reference video, typically requiring fine-tuning for each template which can not only be cumbersome but is restricted to pre-existing motions. VideoComposer~\cite{wang2024videocomposer} and ToonCrafter~\cite{xing2024tooncrafter} incorporate additional inputs, such as depth maps and sketches, to facilitate video generation. These types of input conditions are often difficult to generate, especially for an average user.  DragNUWA~\cite{yin2023dragnuwa} introduced trajectory-based control for video generation, allowing control over both object and camera motion. Subsequent works~\cite{ma2023trailblazer, wang2024motionctrl, wang2024boximator} build on this approach to improve precision and control, though they remain limited to single-image inputs. We propose a method that merges precise and intuitive motion control with the video inbetweening task. 
}

\section{Method}
\label{sec:method}

\cherry{Traditional video inbetweening methods handle simple motions~\cite{reda2022film, niklaus2017adaconv}, while recent diffusion-based methods~\cite{xing2025dynamicrafter, xing2024tooncrafter, feng2024explorative, chen2023seine} boost the generation capability, but provides limited control by strongly relying on model priors and optional text guidance. To facilitate intuitive control, we propose a unified method called \modelname{}, using motion (\eg, trajectories) and content (\eg, masks, guide pixels) guidance to provide precise and user-friendly video inbetweening customization, as shown in Fig.~\ref{fig:pipeline}.

During the training, given the ground truth video clip $X$ with the extracted keyframes \(\{K_f, K_2, \ldots, K_l\}\), we represent the motion control as trajectories consist of sparse RGB points \(\{P_1, P_2, \ldots, P_m\}\) and extract through the proposed \textit{Sparse Motion Generator}; represent the content control as guide pixels \(\{G_1, G_2, \ldots, G_n\}\) and extract through the proposed \textit{Augmented Frame Generator}. 
Sparse points refer to sparse motion trajectory that the model needs to follow and guide pixels refer to specific regions in which the model is instructed to follow certain pixel values. They are integrated together through dual-branch embedders. We choose the DiT-based model as our training backbone due to the long video generation capability. In the remaining section, we provide an overview of the DiT-based model and a detailed discussion of our model design.}

\mt{
% Video interpolation, or inbetweening, typically handles linear motions \cite{reda2022film, niklaus2017adaconv}, while diffusion-based methods \cite{xing2025dynamicrafter, xing2024tooncrafter, feng2024explorative, chen2023seine} rely strongly on model priors and optional text input, which often provides limited control. To enhance intuitive control, we propose a precise, user-friendly method using motion trajectories and region-based controls, including masks and guide pixels.

% As shown in Fig.\ref{fig:pipeline}, our dual-branch design integrates content and motion information in three steps: extracting content from keyframes \(K_f, K_2, \ldots, K_l\), generating sparse point controls \(P_1, P_2, \ldots, P_m\) with the Sparse Motion Generator, and creating guide pixels \(G_1, G_2, \ldots, G_n\) in the Augmented Frame Generator.

% Our curriculum training approach fine-tunes the model in stages, each step adding a new control signal. The following sections provide an overview of the DiT diffusion model structure and a detailed discussion of our training process.

}

\subsection{\cherry{Preliminary}}
% \todoc{introduce basic info for Dit-based Video Gen framework}

% \subsection{DiT structure}

\mt{

% \textbf{Video Diffusion Models}: 
Models like Stable Video Diffusion (SVD), are generative models that extend image diffusion to video by maintaining temporal consistency across frames. Given a noisy video \( X_T \), the model utilizes a conditional 3D-UNet to progressively denoise it to a clean video \( X_0 \) by iteratively applying a denoising function: $X_{t-1} = \epsilon_\theta(X_t, t, c)$, where \( \epsilon_\theta \) represents the learned noise, and $c$ represents the conditions. 

Diffusion Transformer (DiT)~\cite{peebles2023scalable} models combine diffusion-based denoising processes with transformer architectures. Compared to traditional UNet-based models like SVD, DiT leverages a transformer backbone as its core denoiser to model long-range dependencies and global context, which is critical for capturing fine details and significantly improves the versatility and quality of image and video generation.
For training, a diffusion loss is used which measures the mean square error (MSE) between the predicted noise $\hat{\epsilon}$ and the input noise $\epsilon$: $\mathcal{L}_{diff} = || \hat{\epsilon} - \epsilon ||^2_{2}.$
% \cherry{We build our model on top of a text-to-video DiT backbone in a plug-and-play manner.}
Our design is agnostic to the DiT backbone and can be used on different open-source codebases, such as OpenSora.
\begin{figure}[t]
\centering
  \includegraphics[width=0.5\textwidth]{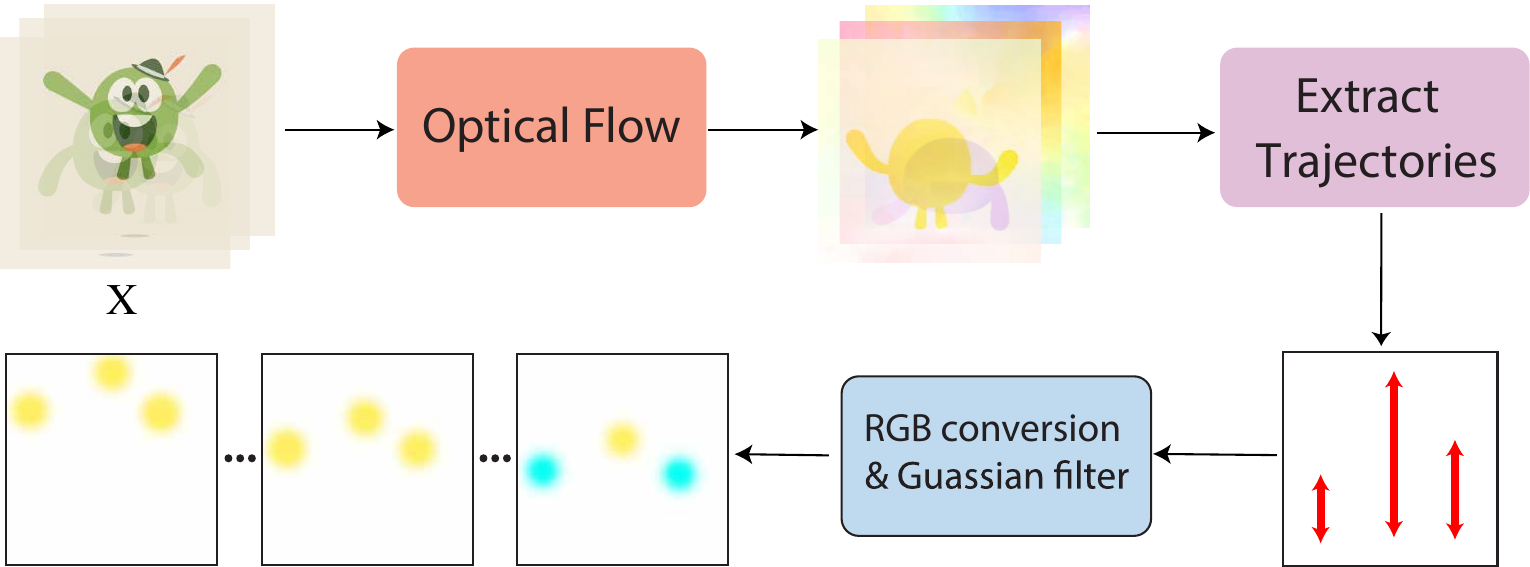}
  \caption{\mt{Structure of Sparse Motion Generator. The input video $X$ is processed through an optical flow generator to extract trajectories. These are then filtered with a Gaussian filter and converted to images to create sparse RGB point controls.}}
  \label{fig:sparse_pipeline}
\end{figure}

% We apply our method to a text-to-video DiT model backbone in a plug-and-play methodology. 
% Fig. \ref{fig:pipeline} shows this process in detail. 
}

\subsection{\cherry{Control Generation}}
% \subsection{Input Pre-Processing}
\label{sec:control_gen}
% \todoc{add two subsections to introduce the sparse motion generator and augmented frame generator.}

\mt{Large-motion interpolation is challenging due to ambiguity, artifacts, and distortions, requiring precise and high-quality control.
% control for quality and control. 
Our project employs two mechanisms: the Sparse Motion Generator, which focuses on key motion paths, and the Augmented Frame Generator, which adds extra visual context. Together, these methods enhance the model’s ability to produce controlled, natural-looking motion.}

\subsubsection{Sparse Motion Generator}
\mt{
The Sparse Motion Generator (Fig.~\ref{fig:sparse_pipeline}) creates motion outputs aligned with the model and the input video $X$. Lacking motion trajectory data, we generate trajectories by extracting a dense optical flow from $X$ and using feature tracking to get motion paths. Since this tracking corresponds to a single pixel, the output is too sparse to be meaningful. To improve interpretability, we expand feature locations using a Gaussian filter similar to~\cite{wang2024motionctrl}, yielding a set of sparse trajectories.

% The first challenge was maintaining the alignment between optical flow and $X$, as DiT models break down the input images into patches.
Due to the patchify module in DiT, which divides input images into patches, aligning the extracted motion trajectories with the $X$ patches is non-trivial.
To address this, we converted trajectories into RGB format, \mt{which means mapping the direction/speed of motion to color space to create a visual representation of the movement}, so that it can follow the same process as the keyframe inputs. 
% \todone{add one sentence explain what does rgb format/point mean}  
The sparse trajectories are thus converted similarly into sparse RGB point controls $\{P_1, P_2, \ldots, P_m\} \in R^{H \times W \times 3}$. While a similar approach is explored in a concurrent work~\cite{zhang2024tora}, we chose a simpler method. Instead of developing a custom VAE for motion, we utilized the DiT’s existing VAE to effectively embed motion, which yielded successful results.

}

\subsubsection{Augmented Frame Generator}
\mt{While motion paths provide effective control over inbetweening, we discovered that the inherent ambiguity of diffusion models, combined with the challenges of interpolating large motions, makes regional control an important enhancement. This approach refines the output, reduces the number of motion paths needed, and supports both static and dynamic regional control. At the same time, we want to avoid overly rigid control, allowing for more natural results. To achieve this, we introduce Augmented Frames. The core concept is to provide the model with a subtle ``nudge" in the right direction, using motion trajectories to guide the output.
To implement this, we extract a region of interest (defined by a mask at inference time) from \(K_f\) and translate it across several frames according to the corresponding trajectory to create frames of guide pixels \(\{G_1, G_2, \ldots, G_n\}\), which are appended to \(K_f\) temporally. For training, we generate masks from motion trajectories using optical-flow segmentation. Further details are available in Fig.~\ref{fig:pipeline}. The ``fox" example in the figure illustrates how we extract the region corresponding to the direction of sparse motion trajectories and append it to the input keyframes.

% In Sec.~\ref{sec:abl_mask} we show this effect in detail. Also in Fig.~\ref{fig:pipeline} we show an implementation of static vs dynamic mask. Similarly in Figs. ~\ref{fig:main_res}, ~\ref{fig:single_res} we show multiple applications. 

One interesting application we identified for this training is the use of ``guide pixels" for providing explicit conditions. 
% Fig.~\ref{fig:exp_res} demonstrates this. 
% After the model learns to interpret partial conditions in \(\{G_1, G_2, \dots, G_n\}\)—where partial input content is provided in otherwise empty frames—we can manually create such inputs and use it to control the generative output, instead of relying on motion trajectories alone to generate the Augmented Frames to provide an additional layer of control.
Once the model learns to interpret guide pixels, we can manually set these guiding regions. Users can specify exactly where the model should place content, such as moving a region from one spot to another. This allows explicit control over the generated frames. The mask and guide pixel controls reduce the need for users to draw extensive trajectories, helping the model accurately identify and track the complete moving object with minimal input.
\cherry{More details and results will be demonstrated in the Experiments section.}
During training, we randomly dropout this content condition by 20\% to support mask-free control.
}

% \yang{shall we mention we also dropout such condition during training to support mask-free control? Or we mention in implementation details that all control modules using dropout training and can be absent during inference. Reviewers might challenge if traj/mask is mandatory.}

\subsection{\cherry{Curriculum Learning for Multi-modal Control with Dual-Branch Encoders}}
\label{sec:cond_diff}
% \todone{shorten this section, to introduce the different levels of learning}

\mt{
To train our model, we utilize a dual-branch encoder structure. First, a set of random keyframes \(\{K_f, K_2, \ldots, K_l\}\) is extracted from \(X\). We always keep the first and last keyframe, and select 0-5 random keyframes in between. From these keyframes, we extract \(\{G_1, G_2, \dots, G_n\}\), for which we use the dense optical flow of \(X\) to create sparse trajectories and optical flow segmentations. The first branch encodes the content information, which includes both \(\{K_f, K_2, \ldots, K_l\}\) and \(\{G_1, G_2, \dots, G_n\}\).

For motion, we extract \(\{P_1, P_2, \ldots, P_m\}\) as discussed above. The second branch encodes this motion information. Both branches have a similar structure. The input (motion or content) is first passed through a frozen VAE to encode it into a latent representation. For content, the latent representation of noise is channel-concatenated with the latent output of conditional images (keyframes and guide pixels). These latent outputs are then passed through Embedders, which first transform the inputs into patches and then funnels the output through a linear layer. The output is again channel-concatenated and passed through a final linear layer before being fed into the transformer denoiser.

%%%%%%%%%%%%%%%%%%%%%%%%
% Furthermore, to embed the motion as a condition, we first performed a dummy experiment. Once we trained our base text-to-video model into an image-to-video model, we used the architecture in Fig ~\ref{fig:pipeline} to train with \(P_1, P_2, \ldots, P_m\). We found that the model quickly learns to ignore the motion input. This analysis is discussed in ~\ref{sec:abl_opt}. To address this issue, we adopted an alternative approach inspired by \cite{yin2023dragnuwa, wang2024motionctrl}. We first trained the model solely with optical flow and then gradually introduced the sparse motion inputs. This phased approach enables the model to better interpret the limited motion information.
%%%%%%%%%%%%%%%%%%%%%%%%

To train our model, we utilize a curriculum training strategy, where we gradually introduce conditional inputs to the model. First, the model is trained only on the image branch to develop a core image interpolation model. 

Afterwards, to embed the motion as a condition, we first performed a dummy experiment. Using the architecture in Fig.~\ref{fig:pipeline} we directly train with \(\{P_1, P_2, \ldots, P_m\}\). From the results we saw that the model quickly learns to ignore the motion input. This analysis is discussed in Sec.~\ref{sec:abl_opt}. To address this issue, we adopted an alternative approach inspired by~\cite{yin2023dragnuwa, wang2024motionctrl}. We first trained the model solely with optical flow and then gradually introduced the sparse motion inputs. This phased approach enables the model to better interpret the limited motion information.
% Next, the model is trained on dense optical flow to recognize motion input, after which we phase in sparse RGB point controls. 
In the last step, we train with guided pixels (\(\{G_1, G_2, \dots, G_n\}\)).

Intuitively, we opted for a two-branch system to separate the two very different conditional inputs. To verify this design choice, we experiment with a single-branch system and share the findings in Sec.~\ref{sec:abl_twobranch}.

}

% We use a sequential method of training, which helps to gradually tune the model with our conditions. We also only provide a single input to the model in a light-weight conditional system which works quite well in incorporating the content and motion information and allows the method to be used with a variety of base models. We now discuss the major steps in our training method.

% \section{Results}
% \label{sec:results}
\section{\cherry{Experiments}}
\label{sec:exps}

% \todone{We show examples of our results in Fig. xx, add some conclusions. Then mention to evaluate our model, we have done xxxxx experiments}

\mt{
To evaluate \modelname{}'s performance, we conduct both quantitative and qualitative assessments across a variety of video sequences and datasets. 
% Visual examples of our results are shown in Fig.~\ref{fig:teaser} and Fig.~\ref{fig:main_res}. 
% These demonstrate the model's ability to generate visually coherent frames that align with user-defined motion paths and masked regions. They show a variety of examples with different combinations of input controls including trajectories, mask and text.
% For example, Fig.~\ref{fig:main_res} top row and Fig.~\ref{fig:teaser} bottom two rows show how masks can be used alongside trajectories to control the interpolation motions. 
For the quantitative assessment, we compare our model’s generative quality and motion control.
}

\textbf{Implementation Details}:
\mt{Our method is applied to a DiT text-to-video diffusion generative model. We use an Adam optimizer with $1\times10^{-4}$ learning rate. Approximately 50k steps are used to train the image-to-video model, 2k steps for optical flow training, 5k for sparse and, 5k for mask input. The entire model, except the VAE and text encoders, is finetuned end-to-end. }

\mt{
% \todoc{check correctness} 
\textit{Automatic Trajectory Generation}: \cherry{For fair comparison with baselines quantitatively,}
% For comparison on large datasets, 
we utilize an automatic trajectory generation method. First, we generate optical flow between the first and last frames. Then, we use feature tracking to create three trajectories, which represent the shortest paths between the two frames.}

\textbf{Baselines}: \mt{Currently, no dedicated methods exist for controllable inbetweening, so we utilize general interpolation techniques for comparison. For this, we select two types of baselines. First, we compare our method with recent diffusion-based video interpolation methods including Explorative Inbetweening of Time and Space (TimeReversal)~\cite{feng2024explorative}, Dynamicrafter~\cite{xing2025dynamicrafter}, and SEINE~\cite{chen2023seine}. We also compare with FILM~\cite{reda2022film}, a non-diffusion method for large motion interpolation.
}

\textbf{Metrics and Datasets}: 
\mt{We use FVD~\cite{unterthiner2019fvd} and LPIPS~\cite{zhang2018unreasonable} for quality comparison.  Additionally, we introduce a ``Motion" metric to evaluate our model's trajectory control. This metric uses the optical flow of the generated output to create trajectory paths corresponding to the input trajectory, and we compute the Fréchet Distance to assess their similarity. We show more details on this in supplementary.
We use DAVIS~\cite{pont20172017} and Objectron~\cite{ahmadyan2021objectron} datasets for general analysis. We also curate a small dataset of 10 videos to analyze the effect of customized motion input.}

\begin{figure*}[t]
\centering
  \includegraphics[width=1\textwidth]{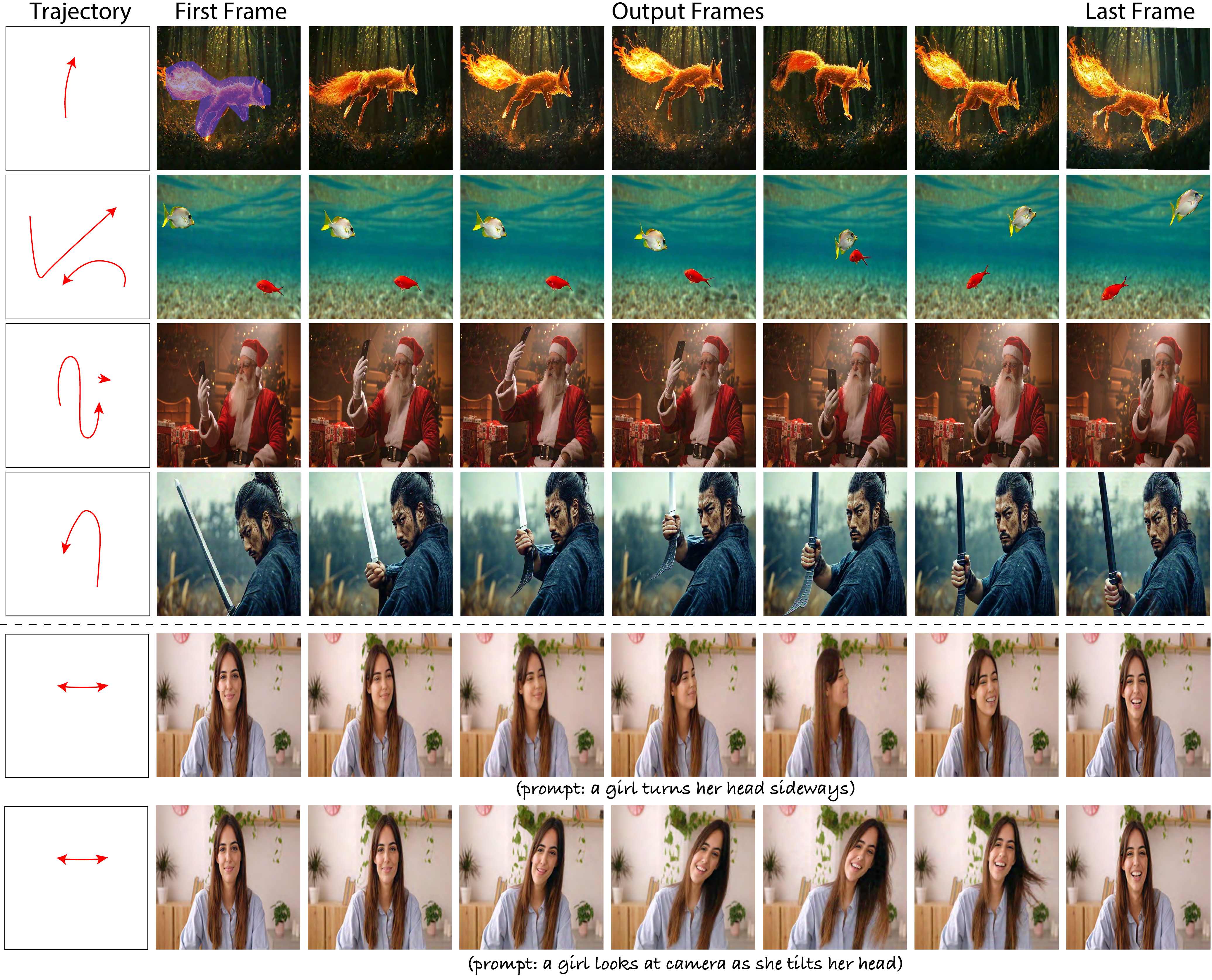}
  \caption{
  % \todoc{Our results. highlight the point, like we can follow the trajectory and other control well, or something. add the motion alignment, replace some of the results to ensure the uniqueness, be consistent with teaser, show text control}
  \mt{Our results. \modelname{} seamlessly integrates motion trajectories and two input frames, enabling smooth transitions. 
  \cherry{Additionally, we present an example using a mask to control the complete object movement in the first row. By further specifying different input prompts, the results are adapted accordingly, as shown in the last two rows. 
  % \yang{I see last three figures show the girl turn her head back which does not follow the trajectory. And leaning back example is not very obvious.}
  }
  % Additionally, we present examples where masks are applied to control movement in specific regions, allowing for targeted manipulation of the animation’s flow. 
  % In cases of ambiguity, further specificity in prompts helps to refine the output. In the last two rows, we showcase an example of different prompts applied to the same motion trajectory, illustrating how prompt variations influence the animation.
  }
  }
  \label{fig:main_res}
\end{figure*}

\begin{figure*}
\centering
  \includegraphics[width=1\textwidth]{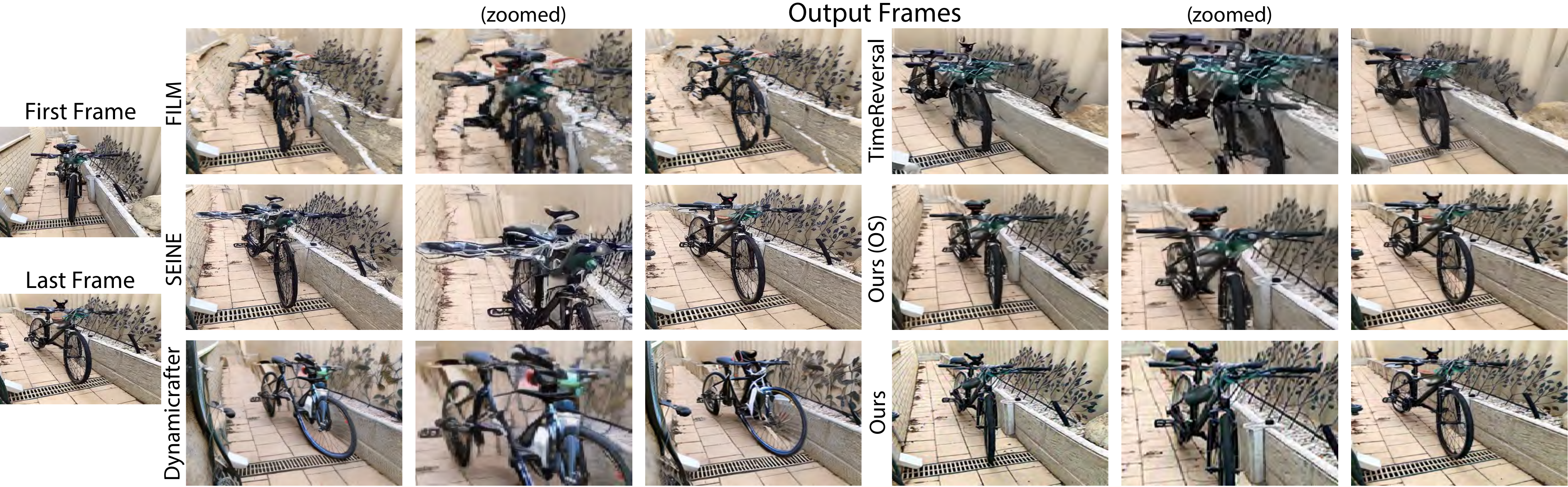}
  \caption{\cherry{Qualitative comparisons. Our model can generate better interpolation results with less artifacts.} 
  % \mt{The interpolation results show that, compared to baselines, our method performs competitively and reduces artifacts and spurious features.}
  % \todoc{Change margins, Replace "Explorative" as "TimeReversal", also add [xx et al. 202x] for each of the method. Check the color change in first example, rotate 90 degrees and show two examples}
  }
  \label{fig:main_qual_intp_all}
\end{figure*}

\subsection{Qualitative Results}
% \todone{directly go to the conclusion about the quality comparisons}

\begin{figure}[t]
\centering
  \includegraphics[width=1\linewidth]{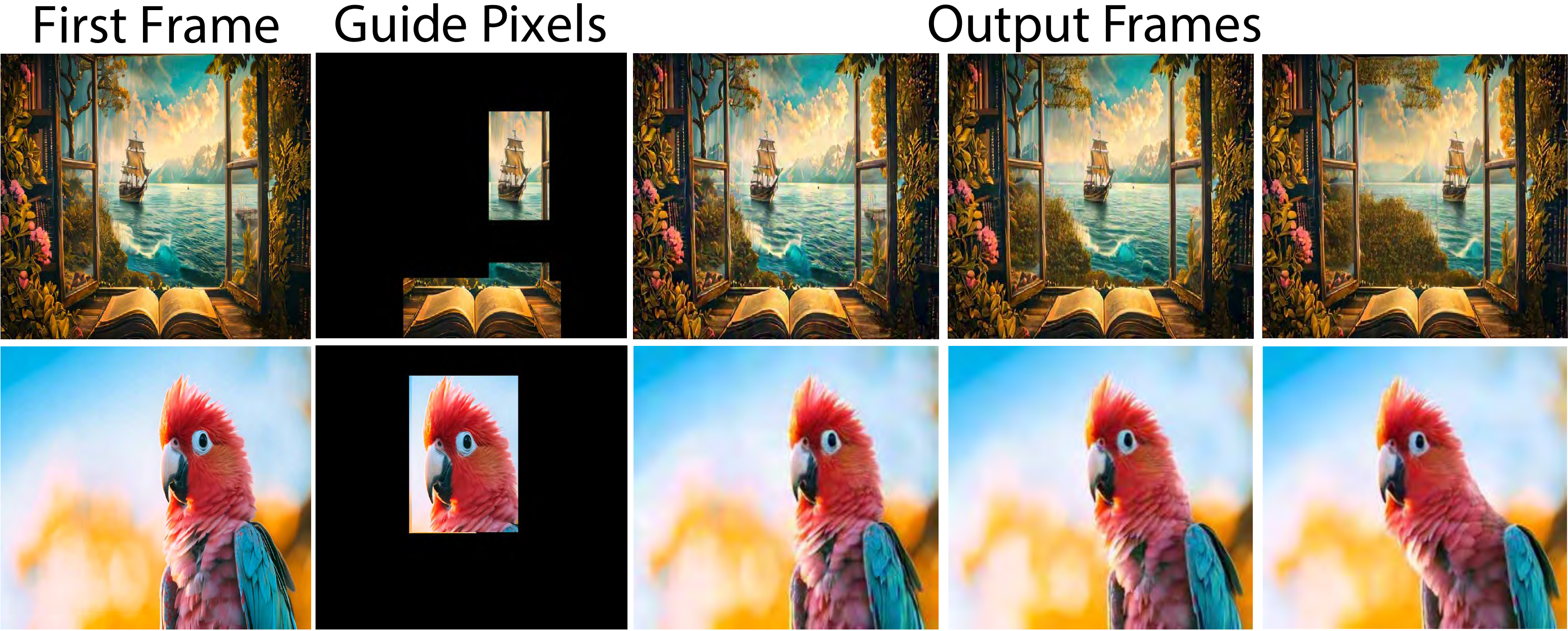}
  \caption{
  \cherry{Example results of taking guide pixels as control. We paste the interested pixels in the target regions for the last frames.}
  % \mt{Examples of providing guide pixels explicitly.}
  }
  \label{fig:exp_res}
\end{figure}

\cherry{As shown in Fig.~\ref{fig:main_res}, we demonstrate}
\mt{
% \todoc{add more justifications such as related work generates artifacts while our model xx, and also refer to concrete examples. make this subsection more informative.}
% Fig.~\ref{fig:main_res} presents a set of qualitative results on diverse inputs, demonstrating 
how our model effectively incorporates both content and motion controls. In the top row, the ``fox" jumps along the specified trajectory, using a single motion vector combined with a mask to define the movement region. The $2^{nd}$ row showcases multi-trajectory results applied to multiple objects, while $3^{rd}$ row illustrates the model's capability to smoothly animate examples like moving a phone and turning a head. 
The last two rows show how effectively the prompt can reduce ambiguity. With the same motion trajectories, and same keyframes, we can adjust the outcome based on provided text.

\cherry{Augmented frames offer an intriguing control where users can paste the interested pixels in the target location as guidance.}
% Augmented Frames offer an intriguing application of utilizing guide pixels for explicit control. 
Fig.~\ref{fig:exp_res} demonstrates example results of manually created augmented frames. By generating these frames and appending them to the end of the video, we are able to produce interesting interpolations.

We also provide a qualitative comparison with baseline methods in Fig.~\ref{fig:main_qual_intp_all}. Our model generates smooth transitions with minimal distortions and artifacts, resulting in natural-looking interpolations. In contrast, FILM~\cite{reda2022film} often morphs keyframes directly, leading to noticeable distortions. Dynamicrafter~\cite{xing2025dynamicrafter} tends to introduce features inconsistent with the keyframes (e.g., changing the appearance of a bike). Both TimeReversal~\cite{feng2024explorative} and SEINE~\cite{chen2023seine} can produce distortions, artifacts, and unsteady motion, which our method effectively minimizes.

}

\begin{table}[h]
\centering
\resizebox{\columnwidth}{!}{%
\begin{tabular}{lccc|ccc|c}
\toprule
\textbf{Method}        & \multicolumn{3}{c}{\textbf{DAVIS}}                      & \multicolumn{3}{c}{\textbf{Objectron}}                & \textbf{Custom}                \\ 
\midrule
                       & \textbf{FVD}\textsubscript{$\downarrow$} & \textbf{LPIPS}\textsubscript{$\downarrow$} & \textbf{Motion}\textsubscript{$\downarrow$} 
                       & \textbf{FVD}\textsubscript{$\downarrow$} & \textbf{LPIPS}\textsubscript{$\downarrow$} & \textbf{Motion}\textsubscript{$\downarrow$} & \textbf{Motion}\textsubscript{$\downarrow$} \\ 
\midrule
\textbf{FILM}          & 2.69                      & \textbf{0.29}                      & 123                   & 3.52                      & \textbf{0.34}             & 113                     & 116                     \\ 
\textbf{SEINE}         & 1.72                      & 0.33                               & 166                   & 2.48                      & 0.38                      & 167                     & 108                     \\ 
\textbf{Dynami}        & 1.79                      & 0.39                               & 226                   & 2.51                      & 0.41                      & 287                     & 104                     \\ 
\textbf{TimeReversal}   & \textbf{1.66}             & 0.36                               & 128                   & 3.61                      & 0.53                      & 132                     & 121                     \\ \midrule
\textbf{Ours}          & \textbf{1.66}             & 0.34                               & 114                   & \textbf{2.37}             & 0.37                      & 116                     & \textbf{75}            \\ 
\textbf{Ours (OS)}     & 1.80                      & 0.35                               & \textbf{106}          & 2.57                      & 0.37                      & \textbf{85}             & 78                      \\ 
\bottomrule
\end{tabular}%
}
\caption{\cherry{Quantitative comparisons with state-of-the-art video inbetweening models. We show our results on different base model architectures and ``OS" represents OpenSora. }
% \todoc{double check the LPIPS}
}
\label{tab:quant_results}
\end{table}

% \begin{table}[]
% \resizebox{0.9\columnwidth}{!}{%
% \begin{tabular}{lcccc}
% % \small
% \toprule
%                & Motion-Study$\uparrow$ & Smooth-Study$\uparrow$ \\ 
% \midrule 
% FILM     & -                   & -       \\
% SEINE            &-          &-          \\
% DynamiCrafter           & -          & -              \\
% TimeReversal           & -           & -           \\
% Ours           & -           & -             \\
% \bottomrule
% \end{tabular}%
% }

% \caption{\mt{User study results for.... \todoc{motion faithfulness and visual quality; change to column figure, use excel/slide to do, complete caption}}}
% \label{tab:tab_user}
% \end{table}

\begin{figure}[t]
\centering
  \includegraphics[width=0.9\linewidth]{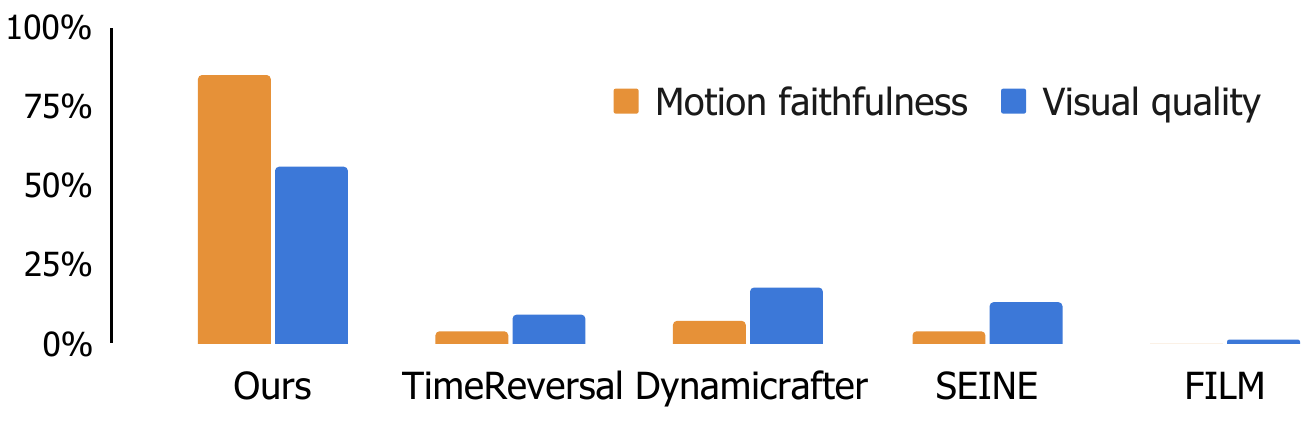}
  \caption{\mt{User study results.
  \cherry{We show the human preference (\%) over two factors: motion faithfulness and visual quality.}
  % : Our model is able to generate high quality videos with strong motion control. 
  % \todoc{motion faithfulness and visual quality; change to column figure, use excel/slide to do, complete caption}
  }}
% \label{tab:tab_user}
  \label{tab:tab_user}
\end{figure}
%priotize mask results especially with complicate close by motion

\subsection{Quantitative Evaluation}
\mt{
Quantitative results are provided in Tab.~\ref{tab:quant_results}. 
\cherry{We randomly sample 100 samples from DAVIS~\cite{pont20172017} and Objectron~\cite{ahmadyan2021objectron} datasets.}
% For DAVIS~\cite{pont20172017} and Objectron~\cite{ahmadyan2021objectron} we use 100 samples each. 
% In FVD, we normalize our results by 700 to facilitate easier interpretation. 
}The results show that our method generates comparable or better results in terms of visual quality compared to the latest interpolation methods, while outperforming related works on motion control by a large margin.
% also providing effective motion control. 
% }

\subsection{User Study}
\cherry{We have conducted user studies to evaluate two factors: \textbf{motion faithfulness} to measure whether the generated results have followed the input motion trajectory, and \textbf{visual quality} to measure whether the generated results are natural and smooth. We randomly select ten results from evaluation sets, and manually design several different motion trajectories for different samples. For each sample, we show results from different methods side by side in a random order and ask participants to choose the best one. More details can be found in the supplementary. We show the results in Fig.~\ref{tab:tab_user}, and demonstrate that}
\mt{% We conducted two user studies: one on motion control and the other on interpolation quality. 
% For 10 diverse examples, we generated outputs using manually defined motion trajectories, while the baselines used default modes. Where applicable, prompts were provided to describe the motion. 
% Results in Fig.~\ref{tab:tab_user} show 
our model effectively follows motion while maintaining high-quality outputs.}

\begin{figure}[t]
\centering
  \includegraphics[width=1\linewidth]{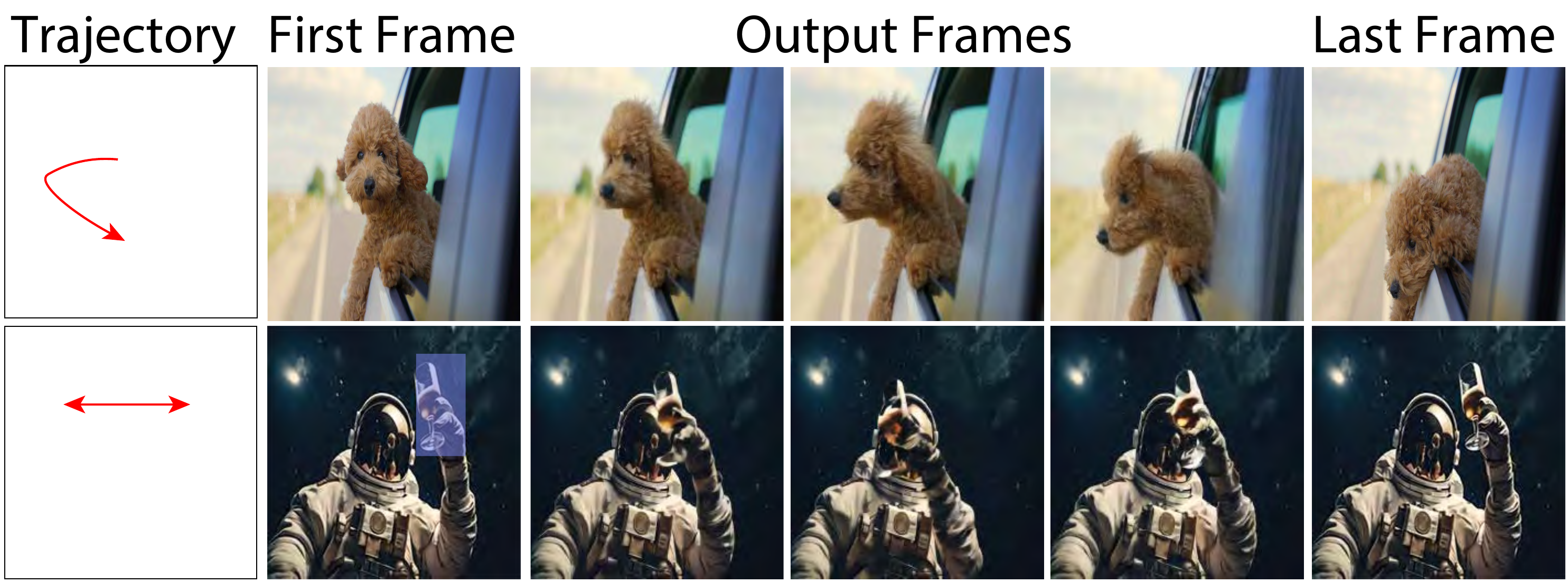}
  \caption{
  % \todone{Generability of our model. draw the same way as teaser, using the same labels} 
  \mt{We verify generalizability of our method by showing results using OpenSora as the base model.}}
  \label{fig:os_res}
\end{figure}

\subsection{Generalizability}
% \todone{explain motivation about this experiment, highlight we are a plug-and-play design, be informative.}
\mt{Our method is designed to work in a backbone-agonistic manner, so it is easy to apply to DiT models with different structures. To verify this generalizability, we show some results with OpenSora (\textbf{OS}) in Fig.~\ref{fig:os_res}. We also use OpenSora to generate quantitative results shared in Tab.~\ref{tab:tab_user}.}

\subsection{Ablation Study}

\begin{figure}[t]
\centering
  \includegraphics[width=1\linewidth]{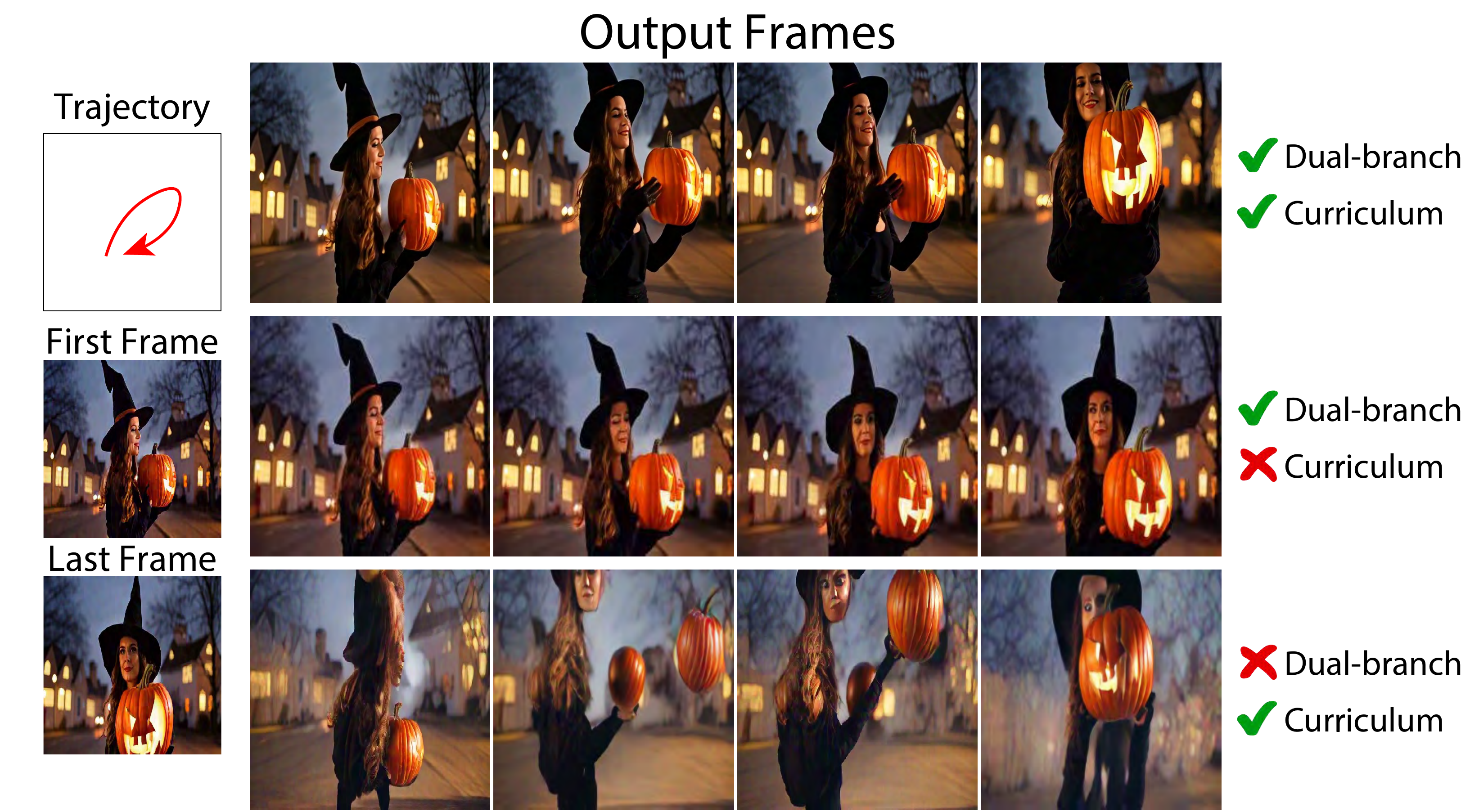}
  % \caption{\todoc{combine with the below curriculum ablation figure, and show different ablation in different rows with ours.} \mt{In the comparison between dual-branch and single-branch implementations, the single-branch model often struggles to effectively follow motion (top) and is prone to noticeable artifacts and content distortions (bottom)}}

  \caption{Results of ablation studies. \mt{
  % The ablation study demonstrates the effectiveness of our design choices. 
  The second row illustrates how the absence of curriculum training prevents the model from accurately recognizing sparse motion trajectories. The last row highlights the importance of dual-branch embedders, showing that using a single branch results in significant distortions.}}
  \label{fig:abl_dual}
\end{figure}

\textbf{The Effectiveness of Curriculum Training.}
\label{sec:abl_opt}
\mt{We initially experimented with \cherry{training directly from sparse motions}.
% using sparse motions directly as input.
However, this approach failed to integrate the motion information, leading the model to focus only on interpolating between the two frames. As shown in Fig.~\ref{fig:abl_dual}, the generated frames effectively bridge the input images but disregard the provided motion cues.}

\noindent\textbf{The Effectiveness of Dual-branch Embedders.}
\label{sec:abl_twobranch}
\mt{In our system, we use two branches to embed content and motion information. In this ablation, we show results using a single branch. Fig.~\ref{fig:abl_dual} demonstrates that significantly more artifacts are present if the two conditions are not separated.}

\begin{figure}[t]
\centering
  \includegraphics[width=1\linewidth]{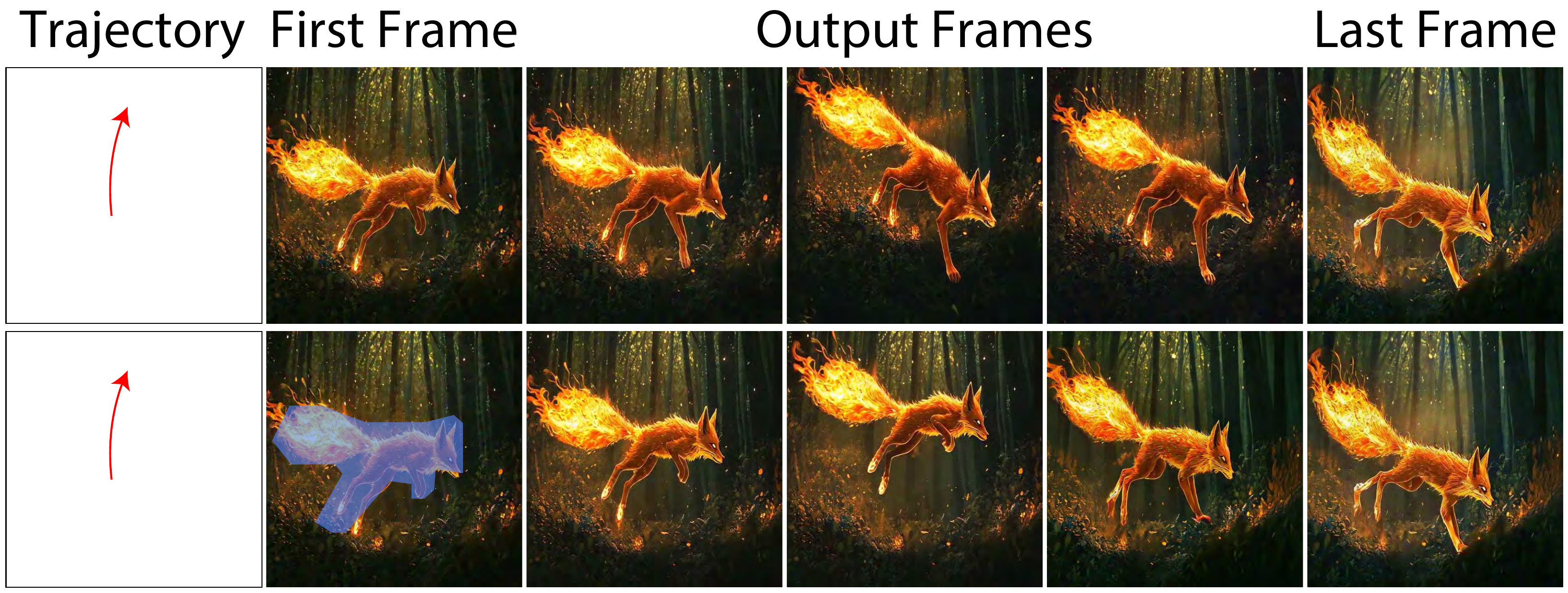}
  \caption{
  % \todone{consistent layout, change or remove the first example, change the showing frames of the second example.}
  Results of using mask control. \mt{It shows how masks enable targeted region movement with fewer sparse input paths. Without a mask (top), movement can be ignored or misdirected (e.g., the fox's back shifts upward), while with a mask (bottom), we achieve the intended jump animation. }}
  \label{fig:abl_mask}
\end{figure}
\noindent\textbf{The Effectiveness of Content Control.}
\label{sec:abl_mask}
\mt{We show the effect of adding masks as opposed to purely using motion vectors as input. Using masks not only improves the reliability and predictability of the results by providing an intuitive control to the user but it also allows for more complicated motions as shown in Fig.~\ref{fig:abl_mask} and also Fig.~\ref{fig:single_res} (top).}

% \todoc{ablation on multikeyframes}
% \subsubsection{Effectiveness of Multi-keyframes}
% \mt{
% While we show results between two images, our method has the capacity to accept multiple images also. We show .... }

\section{Applications}
% \label{sec:exps}
% \todoc{add a section for applications, such as looping, short-to-long video and video generation refinement, etc.}

% \todoc{We show some application here, and due to page limit, we show more applications in supp.}
% \mt{Our method can be applied to a range of potential applications. We highlight several use cases here, and additional examples are provided in the supplementary material.}
\cherry{Our model's versatility supports a wide range of applications and can integrate with existing text-to-video and image-to-video models to refine results with added controls. Here, we showcase several use cases with additional examples available in the supplementary material.
}

\begin{figure}[t]
\centering
  \includegraphics[width=1\linewidth]{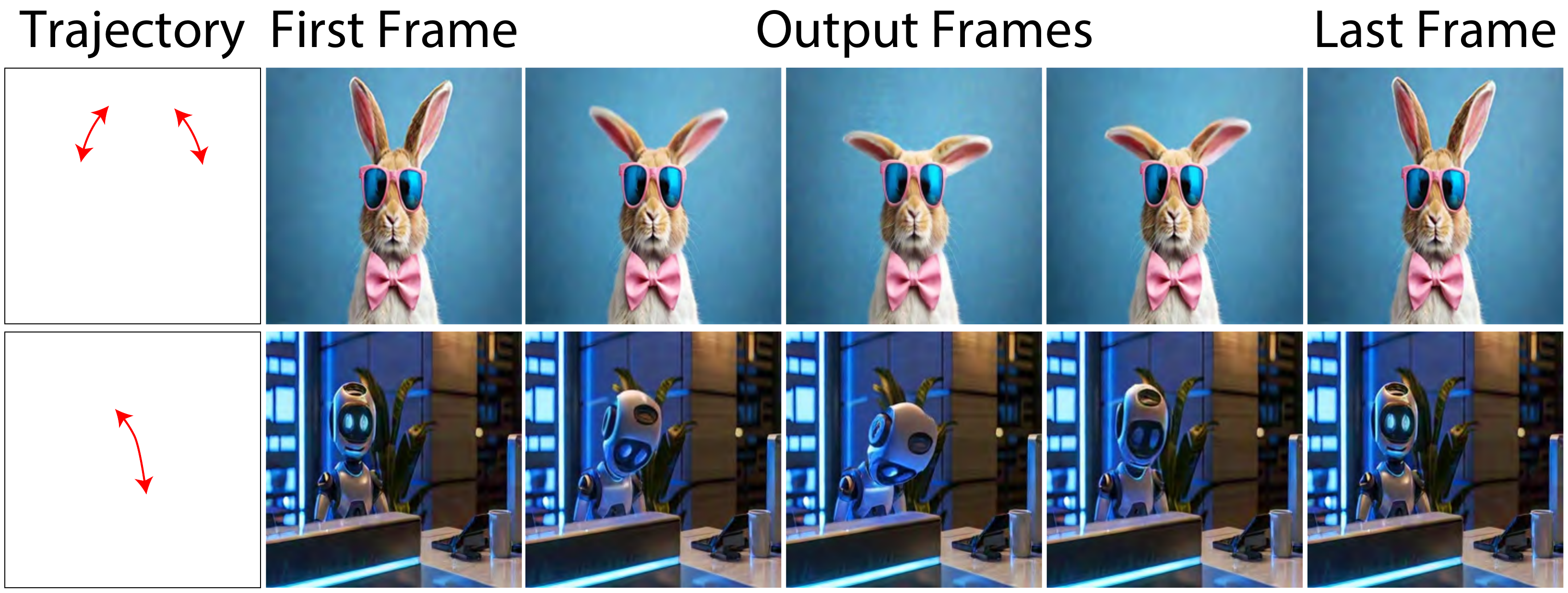}
  \caption{ 
  \mt{Looping video generation. Motion becomes periodic by shortening the trajectory to less than the clip's duration.}
  }
  \label{fig:looping}
\end{figure}
\noindent\textbf{Looping Video Generation.}
\cherry{One application of our model is looping video generation. When the two input frames are identical, our model generates seamless looping videos that precisely follow the input’s motion trajectories, enabling smooth, continuous playback. This capability, illustrated in Fig.~\ref{fig:looping}, is particularly valuable for applications requiring immersive and repetitive animations, such as digital art, virtual environments, and background animations.}
% \mt{When the two input frames are identical, the model generates looping videos that follow the input trajectories 
%  as shown in Fig.~\ref{fig:looping}}.

\begin{figure}[t]
\centering
  \includegraphics[width=1\linewidth]{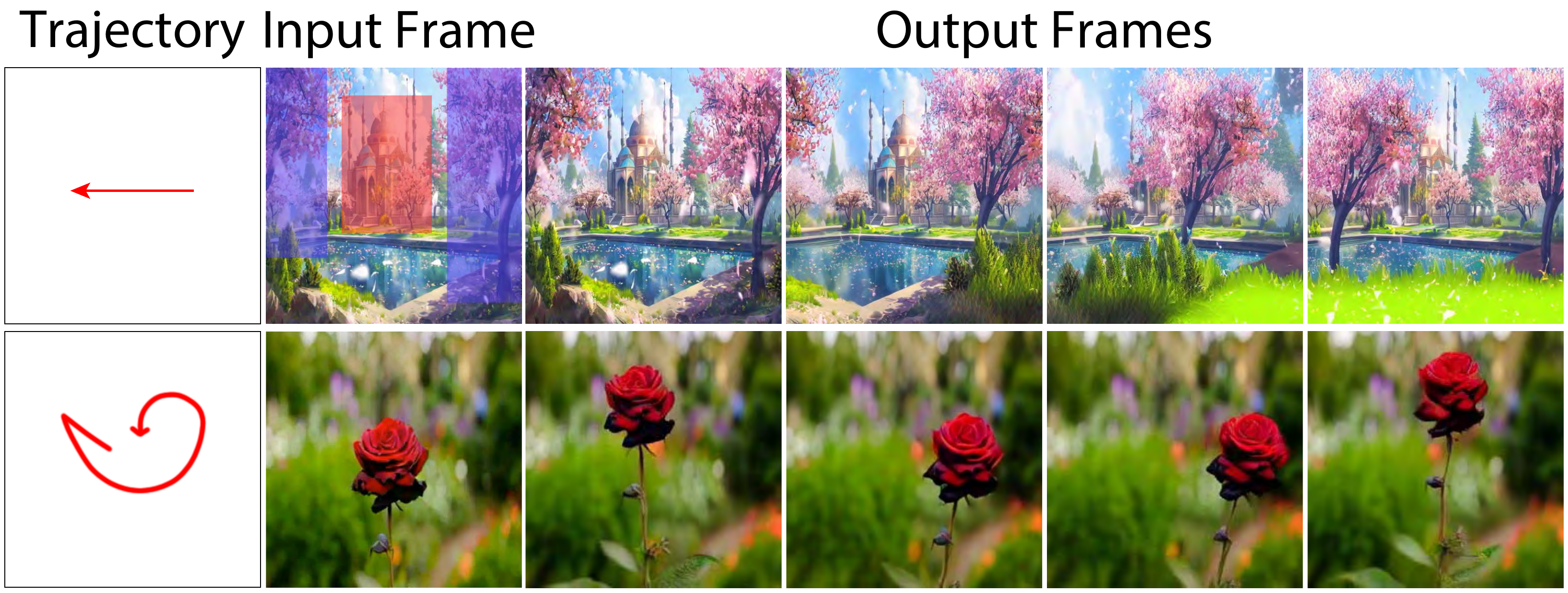}
  \caption{ 
  % \todone{put in application section, replace the first one and use static/dynamic mask colors }
  \mt{Animation results from a single frame. We show the combination of motion trajectory and mask control in the first row and only trajectory control in the second row. Even without training for the single image animation, our model can still generate plausible good results. 
  % \todone{revise label, "input frame"}
  % At the top is a multi mask input, while at the bottom is a trajectory only result.
  }}
  \label{fig:single_res}
\end{figure}

%rethink
\noindent\textbf{Image Animation.}
\cherry{Our model also supports single-image animation, expanding its flexibility beyond inbetweening tasks. Although trained for inbetweening, it can animate a single frame by generating plausible motions, as shown in Fig.~\ref{fig:single_res}. This feature enables creative applications such as bringing static images to life, producing engaging animations from still photos, and enhancing digital storytelling.}
% \mt{Although trained for inbetweening, our method can also handle single-frame input, as shown in Fig.~\ref{fig:single_res}}.

%move to qualitative, as an extra result

% \section{Limitations and Future Work}
% \mt{While our model performs well overall, we have identified several areas for potential improvement in future work. For instance, the trajectories and image content must maintain a certain level of alignment; otherwise, the motion may be overshadowed by the stronger image condition. Additionally, our mask constraint is limited to 2D translation, meaning that 3D movements, such as rotation, cannot currently be captured through masks. Expanding this capability to include 3D transformations would be a valuable direction for more sophisticated control in future versions. }

\section{Conclusion}
\label{sec:conc}
\mt{We introduced \modelname{} a DiT based framework to address the task of controllable inbetweening. Our method is capable of generating high-quality interpolated frames guided by inputs such as keyframes, trajectories and masks. We show the versatility of our method through extensive experiments and applications. 
While our model performs well, we have identified several areas for potential improvement in future work. For instance, the trajectories and image content must maintain a certain level of alignment; otherwise, the motion may be overshadowed by the stronger image condition. Additionally, our mask control is limited to 2D translation, meaning that 3D movements, such as rotation, cannot currently be captured through masks. Expanding this capability to include 3D transformations would be a valuable direction in future versions. }

{
    \small
    \bibliographystyle{ieeenat_fullname}
    \bibliography{main}

\begin{thebibliography}{50}
\providecommand{\natexlab}[1]{#1}
\providecommand{\url}[1]{\texttt{#1}}
\expandafter\ifx\csname urlstyle\endcsname\relax
  \providecommand{\doi}[1]{doi: #1}\else
  \providecommand{\doi}{doi: \begingroup \urlstyle{rm}\Url}\fi

\bibitem[Ahmadyan et~al.(2021)Ahmadyan, Zhang, Ablavatski, Wei, and Grundmann]{ahmadyan2021objectron}
Adel Ahmadyan, Liangkai Zhang, Artsiom Ablavatski, Jianing Wei, and Matthias Grundmann.
\newblock Objectron: A large scale dataset of object-centric videos in the wild with pose annotations.
\newblock In \emph{Proceedings of the IEEE/CVF conference on computer vision and pattern recognition}, pages 7822--7831, 2021.

\bibitem[Blattmann et~al.(2023{\natexlab{a}})Blattmann, Dockhorn, Kulal, Mendelevitch, Kilian, Lorenz, Levi, English, Voleti, Letts, et~al.]{blattmann2023stable}
Andreas Blattmann, Tim Dockhorn, Sumith Kulal, Daniel Mendelevitch, Maciej Kilian, Dominik Lorenz, Yam Levi, Zion English, Vikram Voleti, Adam Letts, et~al.
\newblock Stable video diffusion: Scaling latent video diffusion models to large datasets.
\newblock \emph{arXiv preprint arXiv:2311.15127}, 2023{\natexlab{a}}.

\bibitem[Blattmann et~al.(2023{\natexlab{b}})Blattmann, Rombach, Ling, Dockhorn, Kim, Fidler, and Kreis]{blattmann2023align}
Andreas Blattmann, Robin Rombach, Huan Ling, Tim Dockhorn, Seung~Wook Kim, Sanja Fidler, and Karsten Kreis.
\newblock Align your latents: High-resolution video synthesis with latent diffusion models.
\newblock In \emph{Proceedings of the IEEE/CVF Conference on Computer Vision and Pattern Recognition}, pages 22563--22575, 2023{\natexlab{b}}.

\bibitem[Chen et~al.(2023)Chen, Wang, Zhang, Zhuang, Ma, Yu, Wang, Lin, Qiao, and Liu]{chen2023seine}
Xinyuan Chen, Yaohui Wang, Lingjun Zhang, Shaobin Zhuang, Xin Ma, Jiashuo Yu, Yali Wang, Dahua Lin, Yu Qiao, and Ziwei Liu.
\newblock Seine: Short-to-long video diffusion model for generative transition and prediction.
\newblock In \emph{The Twelfth International Conference on Learning Representations}, 2023.

\bibitem[Choi et~al.(2000)Choi, Lee, and Ko]{choi2000fruc}
Byung{-}Tae Choi, Sung{-}Hee Lee, and Sung{-}Jea Ko.
\newblock New frame rate up-conversion using bi-directional motion estimation.
\newblock \emph{{IEEE} Trans. Consumer Electron.}, 46\penalty0 (3):\penalty0 603--609, 2000.

\bibitem[Feng et~al.(2024)Feng, Ding, Xia, Niklaus, Abrevaya, Black, and Zhang]{feng2024explorative}
Haiwen Feng, Zheng Ding, Zhihao Xia, Simon Niklaus, Victoria Abrevaya, Michael~J Black, and Xuaner Zhang.
\newblock Explorative inbetweening of time and space.
\newblock \emph{arXiv preprint arXiv:2403.14611}, 2024.

\bibitem[Girdhar et~al.(2023)Girdhar, Singh, Brown, Duval, Azadi, Rambhatla, Shah, Yin, Parikh, and Misra]{girdhar2023emu}
Rohit Girdhar, Mannat Singh, Andrew Brown, Quentin Duval, Samaneh Azadi, Sai~Saketh Rambhatla, Akbar Shah, Xi Yin, Devi Parikh, and Ishan Misra.
\newblock Emu video: Factorizing text-to-video generation by explicit image conditioning.
\newblock \emph{arXiv preprint arXiv:2311.10709}, 2023.

\bibitem[Ha et~al.(2004)Ha, Lee, and Kim]{ha2004moco}
Taehyeun Ha, Seongjoo Lee, and Jaeseok Kim.
\newblock Motion compensated frame interpolation by new block-based motion estimation algorithm.
\newblock \emph{{IEEE} Trans. Consumer Electron.}, 50\penalty0 (2):\penalty0 752--759, 2004.

\bibitem[Ho et~al.(2022{\natexlab{a}})Ho, Chan, Saharia, Whang, Gao, Gritsenko, Kingma, Poole, Norouzi, Fleet, et~al.]{ho2022imagen}
Jonathan Ho, William Chan, Chitwan Saharia, Jay Whang, Ruiqi Gao, Alexey Gritsenko, Diederik~P Kingma, Ben Poole, Mohammad Norouzi, David~J Fleet, et~al.
\newblock Imagen video: High definition video generation with diffusion models.
\newblock \emph{arXiv preprint arXiv:2210.02303}, 2022{\natexlab{a}}.

\bibitem[Ho et~al.(2022{\natexlab{b}})Ho, Salimans, Gritsenko, Chan, Norouzi, and Fleet]{ho2022video}
Jonathan Ho, Tim Salimans, Alexey Gritsenko, William Chan, Mohammad Norouzi, and David~J Fleet.
\newblock Video diffusion models.
\newblock \emph{Advances in Neural Information Processing Systems}, 35:\penalty0 8633--8646, 2022{\natexlab{b}}.

\bibitem[Hong et~al.(2023)Hong, Seo, Shin, Hong, and Kim]{hong2023large}
Susung Hong, Junyoung Seo, Heeseong Shin, Sunghwan Hong, and Seungryong Kim.
\newblock Large language models are frame-level directors for zero-shot text-to-video generation.
\newblock In \emph{First Workshop on Controllable Video Generation@ ICML24}, 2023.

\bibitem[Hong et~al.(2022)Hong, Ding, Zheng, Liu, and Tang]{hong2022cogvideo}
Wenyi Hong, Ming Ding, Wendi Zheng, Xinghan Liu, and Jie Tang.
\newblock Cogvideo: Large-scale pretraining for text-to-video generation via transformers, 2022.

\bibitem[Kwon et~al.(2024)Kwon, Oh, Zhou, Liu, Lee, Cai, Liu, Liu, and Uh]{kwon2024harivo}
Mingi Kwon, Seoung~Wug Oh, Yang Zhou, Difan Liu, Joon-Young Lee, Haoran Cai, Baqiao Liu, Feng Liu, and Youngjung Uh.
\newblock Harivo: Harnessing text-to-image models for video generation.
\newblock \emph{arXiv preprint arXiv:2410.07763}, 2024.

\bibitem[Li et~al.(2023)Li, Chu, Wu, Yuan, Liu, Zhang, Li, Feng, Ding, and Wang]{li2023videogen}
Xin Li, Wenqing Chu, Ye Wu, Weihang Yuan, Fanglong Liu, Qi Zhang, Fu Li, Haocheng Feng, Errui Ding, and Jingdong Wang.
\newblock Videogen: A reference-guided latent diffusion approach for high definition text-to-video generation.
\newblock \emph{arXiv preprint arXiv:2309.00398}, 2023.

\bibitem[Ma et~al.(2023)Ma, Lewis, and Kleijn]{ma2023trailblazer}
Wan-Duo~Kurt Ma, John~P Lewis, and W~Bastiaan Kleijn.
\newblock Trailblazer: Trajectory control for diffusion-based video generation.
\newblock \emph{arXiv preprint arXiv:2401.00896}, 2023.

\bibitem[Meyer et~al.(2015)Meyer, Wang, Zimmer, Grosse, and Sorkine{-}Hornung]{meyer2015phasebased}
Simone Meyer, Oliver Wang, Henning Zimmer, Max Grosse, and Alexander Sorkine{-}Hornung.
\newblock Phase-based frame interpolation for video.
\newblock In \emph{{CVPR}}, pages 1410--1418. {IEEE} Computer Society, 2015.

\bibitem[Meyer et~al.(2018)Meyer, Cornill{\`{e}}re, Djelouah, Schroers, and Gross]{meyer2018colorprop}
Simone Meyer, Victor Cornill{\`{e}}re, Abdelaziz Djelouah, Christopher Schroers, and Markus~H. Gross.
\newblock Deep video color propagation.
\newblock In \emph{{BMVC}}, page 128. {BMVA} Press, 2018.

\bibitem[Niklaus and Liu(2018)]{niklaus2018ctxsyn}
Simon Niklaus and Feng Liu.
\newblock Context-aware synthesis for video frame interpolation.
\newblock In \emph{{CVPR}}, pages 1701--1710. Computer Vision Foundation / {IEEE} Computer Society, 2018.

\bibitem[Niklaus and Liu(2020)]{niklaus2020softsplat}
Simon Niklaus and Feng Liu.
\newblock Softmax splatting for video frame interpolation.
\newblock In \emph{{CVPR}}, pages 5436--5445. Computer Vision Foundation / {IEEE}, 2020.

\bibitem[Niklaus et~al.(2017{\natexlab{a}})Niklaus, Mai, and Liu]{niklaus2017adaconv}
Simon Niklaus, Long Mai, and Feng Liu.
\newblock Video frame interpolation via adaptive convolution.
\newblock In \emph{Proceedings of the IEEE conference on computer vision and pattern recognition}, pages 670--679, 2017{\natexlab{a}}.

\bibitem[Niklaus et~al.(2017{\natexlab{b}})Niklaus, Mai, and Liu]{niklaus2017sepconv}
Simon Niklaus, Long Mai, and Feng Liu.
\newblock Video frame interpolation via adaptive separable convolution.
\newblock In \emph{{ICCV}}, pages 261--270. {IEEE} Computer Society, 2017{\natexlab{b}}.

\bibitem[Niklaus et~al.(2023)Niklaus, Hu, and Chen]{niklaus2023splatsyn}
Simon Niklaus, Ping Hu, and Jiawen Chen.
\newblock Splatting-based synthesis for video frame interpolation.
\newblock In \emph{{WACV}}, pages 713--723. {IEEE}, 2023.

\bibitem[Peebles and Xie(2023)]{peebles2023scalable}
William Peebles and Saining Xie.
\newblock Scalable diffusion models with transformers.
\newblock In \emph{Proceedings of the IEEE/CVF International Conference on Computer Vision}, pages 4195--4205, 2023.

\bibitem[Pont-Tuset et~al.(2017)Pont-Tuset, Perazzi, Caelles, Arbel{\'a}ez, Sorkine-Hornung, and Van~Gool]{pont20172017}
Jordi Pont-Tuset, Federico Perazzi, Sergi Caelles, Pablo Arbel{\'a}ez, Alex Sorkine-Hornung, and Luc Van~Gool.
\newblock The 2017 davis challenge on video object segmentation.
\newblock \emph{arXiv preprint arXiv:1704.00675}, 2017.

\bibitem[Ranzato et~al.(2016)Ranzato, Szlam, Bruna, Mathieu, Collobert, and Chopra]{ranzato2016video}
MarcAurelio Ranzato, Arthur Szlam, Joan Bruna, Michael Mathieu, Ronan Collobert, and Sumit Chopra.
\newblock Video (language) modeling: a baseline for generative models of natural videos, 2016.

\bibitem[Reda et~al.(2022)Reda, Kontkanen, Tabellion, Sun, Pantofaru, and Curless]{reda2022film}
Fitsum Reda, Janne Kontkanen, Eric Tabellion, Deqing Sun, Caroline Pantofaru, and Brian Curless.
\newblock Film: Frame interpolation for large motion.
\newblock In \emph{European Conference on Computer Vision}, pages 250--266. Springer, 2022.

\bibitem[Ren et~al.(2024)Ren, Zhou, Yang, Shi, Liu, Liu, Kwon, and Shrivastava]{ren2024customize}
Yixuan Ren, Yang Zhou, Jimei Yang, Jing Shi, Difan Liu, Feng Liu, Mingi Kwon, and Abhinav Shrivastava.
\newblock Customize-a-video: One-shot motion customization of text-to-video diffusion models.
\newblock \emph{arXiv preprint arXiv:2402.14780}, 2024.

\bibitem[Rombach et~al.(2022)Rombach, Blattmann, Lorenz, Esser, and Ommer]{rombach2022high}
Robin Rombach, Andreas Blattmann, Dominik Lorenz, Patrick Esser, and Bj{\"o}rn Ommer.
\newblock High-resolution image synthesis with latent diffusion models.
\newblock In \emph{Proceedings of the IEEE/CVF conference on computer vision and pattern recognition}, pages 10684--10695, 2022.

\bibitem[Saito et~al.(2017)Saito, Matsumoto, and Saito]{saito2017temporal}
Masaki Saito, Eiichi Matsumoto, and Shunta Saito.
\newblock Temporal generative adversarial nets with singular value clipping, 2017.

\bibitem[Shen et~al.(2023)Shen, Li, and Elhoseiny]{shen2023mostganv}
Xiaoqian Shen, Xiang Li, and Mohamed Elhoseiny.
\newblock Mostgan-v: Video generation with temporal motion styles, 2023.

\bibitem[Singer et~al.(2022)Singer, Polyak, Hayes, Yin, An, Zhang, Hu, Yang, Ashual, Gafni, et~al.]{singer2022make}
Uriel Singer, Adam Polyak, Thomas Hayes, Xi Yin, Jie An, Songyang Zhang, Qiyuan Hu, Harry Yang, Oron Ashual, Oran Gafni, et~al.
\newblock Make-a-video: Text-to-video generation without text-video data.
\newblock \emph{arXiv preprint arXiv:2209.14792}, 2022.

\bibitem[Siyao et~al.(2021)Siyao, Zhao, Yu, Sun, Metaxas, Loy, and Liu]{siyao2021animeinterp}
Li Siyao, Shiyu Zhao, Weijiang Yu, Wenxiu Sun, Dimitris~N. Metaxas, Chen~Change Loy, and Ziwei Liu.
\newblock Deep animation video interpolation in the wild.
\newblock In \emph{{CVPR}}, pages 6587--6595. Computer Vision Foundation / {IEEE}, 2021.

\bibitem[Srivastava et~al.(2016)Srivastava, Mansimov, and Salakhutdinov]{srivastava2016unsupervised}
Nitish Srivastava, Elman Mansimov, and Ruslan Salakhutdinov.
\newblock Unsupervised learning of video representations using lstms, 2016.

\bibitem[Tulyakov et~al.(2017)Tulyakov, Liu, Yang, and Kautz]{tulyakov2017mocogan}
Sergey Tulyakov, Ming-Yu Liu, Xiaodong Yang, and Jan Kautz.
\newblock Mocogan: Decomposing motion and content for video generation, 2017.

\bibitem[Unterthiner et~al.(2019)Unterthiner, van Steenkiste, Kurach, Marinier, Michalski, and Gelly]{unterthiner2019fvd}
Thomas Unterthiner, Sjoerd van Steenkiste, Karol Kurach, Rapha{\"e}l Marinier, Marcin Michalski, and Sylvain Gelly.
\newblock Fvd: A new metric for video generation.
\newblock In \emph{ICLR 2019 Workshop DeepGenStruct}, 2019.

\bibitem[Wang et~al.(2024{\natexlab{a}})Wang, Zhang, Zou, Zeng, Wei, Yuan, and Li]{wang2024boximator}
Jiawei Wang, Yuchen Zhang, Jiaxin Zou, Yan Zeng, Guoqiang Wei, Liping Yuan, and Hang Li.
\newblock Boximator: Generating rich and controllable motions for video synthesis.
\newblock \emph{arXiv preprint arXiv:2402.01566}, 2024{\natexlab{a}}.

\bibitem[Wang et~al.(2024{\natexlab{b}})Wang, Yuan, Zhang, Chen, Wang, Zhang, Shen, Zhao, and Zhou]{wang2024videocomposer}
Xiang Wang, Hangjie Yuan, Shiwei Zhang, Dayou Chen, Jiuniu Wang, Yingya Zhang, Yujun Shen, Deli Zhao, and Jingren Zhou.
\newblock Videocomposer: Compositional video synthesis with motion controllability.
\newblock \emph{Advances in Neural Information Processing Systems}, 36, 2024{\natexlab{b}}.

\bibitem[Wang et~al.(2024{\natexlab{c}})Wang, Bao, Weng, Feng, Yin, Yang, Zhang, Dai, Zhao, Wang, et~al.]{wang2024microcinema}
Yanhui Wang, Jianmin Bao, Wenming Weng, Ruoyu Feng, Dacheng Yin, Tao Yang, Jingxu Zhang, Qi Dai, Zhiyuan Zhao, Chunyu Wang, et~al.
\newblock Microcinema: A divide-and-conquer approach for text-to-video generation.
\newblock In \emph{Proceedings of the IEEE/CVF Conference on Computer Vision and Pattern Recognition}, pages 8414--8424, 2024{\natexlab{c}}.

\bibitem[Wang et~al.(2024{\natexlab{d}})Wang, Yuan, Wang, Li, Chen, Xia, Luo, and Shan]{wang2024motionctrl}
Zhouxia Wang, Ziyang Yuan, Xintao Wang, Yaowei Li, Tianshui Chen, Menghan Xia, Ping Luo, and Ying Shan.
\newblock Motionctrl: A unified and flexible motion controller for video generation.
\newblock In \emph{ACM SIGGRAPH 2024 Conference Papers}, pages 1--11, 2024{\natexlab{d}}.

\bibitem[Wu et~al.(2023)Wu, Ge, Wang, Lei, Gu, Shi, Hsu, Shan, Qie, and Shou]{wu2023tune}
Jay~Zhangjie Wu, Yixiao Ge, Xintao Wang, Stan~Weixian Lei, Yuchao Gu, Yufei Shi, Wynne Hsu, Ying Shan, Xiaohu Qie, and Mike~Zheng Shou.
\newblock Tune-a-video: One-shot tuning of image diffusion models for text-to-video generation.
\newblock In \emph{Proceedings of the IEEE/CVF International Conference on Computer Vision}, pages 7623--7633, 2023.

\bibitem[Xing et~al.(2024)Xing, Liu, Xia, Zhang, Wang, Shan, and Wong]{xing2024tooncrafter}
Jinbo Xing, Hanyuan Liu, Menghan Xia, Yong Zhang, Xintao Wang, Ying Shan, and Tien-Tsin Wong.
\newblock Tooncrafter: Generative cartoon interpolation.
\newblock \emph{arXiv preprint arXiv:2405.17933}, 2024.

\bibitem[Xing et~al.(2025)Xing, Xia, Zhang, Chen, Yu, Liu, Liu, Wang, Shan, and Wong]{xing2025dynamicrafter}
Jinbo Xing, Menghan Xia, Yong Zhang, Haoxin Chen, Wangbo Yu, Hanyuan Liu, Gongye Liu, Xintao Wang, Ying Shan, and Tien-Tsin Wong.
\newblock Dynamicrafter: Animating open-domain images with video diffusion priors.
\newblock In \emph{European Conference on Computer Vision}, pages 399--417. Springer, 2025.

\bibitem[Yan et~al.(2021)Yan, Zhang, Abbeel, and Srinivas]{yan2021videogpt}
Wilson Yan, Yunzhi Zhang, Pieter Abbeel, and Aravind Srinivas.
\newblock Videogpt: Video generation using vq-vae and transformers, 2021.

\bibitem[Yin et~al.(2023)Yin, Wu, Liang, Shi, Li, Ming, and Duan]{yin2023dragnuwa}
Shengming Yin, Chenfei Wu, Jian Liang, Jie Shi, Houqiang Li, Gong Ming, and Nan Duan.
\newblock Dragnuwa: Fine-grained control in video generation by integrating text, image, and trajectory.
\newblock \emph{arXiv preprint arXiv:2308.08089}, 2023.

\bibitem[Yu et~al.(2023)Yu, Cheng, Sohn, Lezama, Zhang, Chang, Hauptmann, Yang, Hao, Essa, and Jiang]{yu2023magvit}
Lijun Yu, Yong Cheng, Kihyuk Sohn, José Lezama, Han Zhang, Huiwen Chang, Alexander~G. Hauptmann, Ming-Hsuan Yang, Yuan Hao, Irfan Essa, and Lu Jiang.
\newblock Magvit: Masked generative video transformer, 2023.

\bibitem[Zeng et~al.(2024)Zeng, Wei, Zheng, Zou, Wei, Zhang, and Li]{zeng2024make}
Yan Zeng, Guoqiang Wei, Jiani Zheng, Jiaxin Zou, Yang Wei, Yuchen Zhang, and Hang Li.
\newblock Make pixels dance: High-dynamic video generation.
\newblock In \emph{Proceedings of the IEEE/CVF Conference on Computer Vision and Pattern Recognition}, pages 8850--8860, 2024.

\bibitem[Zhang et~al.(2018)Zhang, Isola, Efros, Shechtman, and Wang]{zhang2018unreasonable}
Richard Zhang, Phillip Isola, Alexei~A Efros, Eli Shechtman, and Oliver Wang.
\newblock The unreasonable effectiveness of deep features as a perceptual metric.
\newblock In \emph{Proceedings of the IEEE conference on computer vision and pattern recognition}, pages 586--595, 2018.

\bibitem[Zhang et~al.(2024)Zhang, Liao, Li, Qin, and Wang]{zhang2024tora}
Zhenghao Zhang, Junchao Liao, Menghao Li, Long Qin, and Weizhi Wang.
\newblock Tora: Trajectory-oriented diffusion transformer for video generation.
\newblock \emph{arXiv preprint arXiv:2407.21705}, 2024.

\bibitem[Zhao et~al.(2025)Zhao, Gu, Wu, Zhang, Liu, Wu, Keppo, and Shou]{zhao2025motiondirector}
Rui Zhao, Yuchao Gu, Jay~Zhangjie Wu, David~Junhao Zhang, Jia-Wei Liu, Weijia Wu, Jussi Keppo, and Mike~Zheng Shou.
\newblock Motiondirector: Motion customization of text-to-video diffusion models.
\newblock In \emph{European Conference on Computer Vision}, pages 273--290. Springer, 2025.

\bibitem[Zhou et~al.(2022)Zhou, Yang, Li, Saito, Aneja, and Kalogerakis]{zhou2022audio}
Yang Zhou, Jimei Yang, Dingzeyu Li, Jun Saito, Deepali Aneja, and Evangelos Kalogerakis.
\newblock Audio-driven neural gesture reenactment with video motion graphs.
\newblock In \emph{Proceedings of the IEEE/CVF conference on computer vision and pattern recognition}, pages 3418--3428, 2022.

\end{thebibliography}
}

% \input{figs/main_results_supp}

\end{document}